\documentclass[preprint,5p,times,twocolumn]{elsarticle}

\usepackage{amssymb}
\usepackage{amsmath}
\usepackage{multirow}
\usepackage{array}
\usepackage{longtable}
\usepackage{tabularx} 
\usepackage{lipsum}   
\usepackage{amssymb} 
\usepackage{booktabs}      
\usepackage{pifont} 
\newcommand{\cmark}{\ding{51}} 
\newcommand{\xmark}{\ding{55}} 

\usepackage{xcolor}  
\usepackage{xurl}

\usepackage{xr}

\journal{ACM Computing Surveys}

\begin{document}

\begin{frontmatter}



\title{Few-Shot Learning in Video and 3D Object Detection: A Survey} 


\author[inst1]{Md
Meftahul Ferdaus}
\affiliation[inst1]{organization={Canizaro Livingston Gulf States Center for Environmental Informatics},
            addressline={University of New Orleans}, 
            city={New Orleans},
            postcode={70148}, 
            state={Louisiana},
            country={USA}}

\author[inst2]{Kendall N. Niles}
\author[inst2]{Joe Tom}
\affiliation[inst2]{organization={US Army Corps of Engineers},
            city={Vicksburg},
            postcode={39183}, 
            state={Mississippi},
            country={USA}}

\author[inst1]{Mahdi Abdelguerfi}

\author[inst3]{Elias Ioup}
\affiliation[inst3]{organization={Center for Geospatial Sciences, Naval Research Laboratory},
            addressline={Stennis Space Center}, 
            city={Hancock County},
            postcode={39529}, 
            state={Mississippi},
            country={USA}}


\begin{abstract}
Few-shot learning (FSL) enables object detection models to recognize novel classes given only a few annotated examples, thereby reducing expensive manual data labeling. This survey examines recent FSL advances for video and 3D object detection. For video, FSL is especially valuable since annotating objects across frames is more laborious than for static images. By propagating information across frames, techniques like tube proposals and temporal matching networks can detect new classes from a couple examples, efficiently leveraging spatiotemporal structure. FSL for 3D detection from LiDAR or depth data faces challenges like sparsity and lack of texture. Solutions integrate FSL with specialized point cloud networks and losses tailored for class imbalance. Few-shot 3D detection enables practical autonomous driving deployment by minimizing costly 3D annotation needs. Core issues in both domains include balancing generalization and overfitting, integrating prototype matching, and handling data modality properties. In summary, FSL shows promise for reducing annotation requirements and enabling real-world video, 3D, and other applications by efficiently leveraging information across feature, temporal, and data modalities. By comprehensively surveying recent advancements, this paper illuminates FSL's potential to minimize supervision needs and enable deployment across video, 3D, and other real-world applications.
\end{abstract}



\begin{keyword}
Data Scarcity \sep Few-Shot Learning \sep Modern Tracking Systems \sep Video \sep 3D Object Detection



\end{keyword}

\end{frontmatter}


\section{Introduction}\label{sec:introduction}
Object detection is a fundamental task in computer vision that involves locating and classifying objects belonging to predefined categories in images or video frames \cite{lin2017focal}. Over the years, deep convolutional neural networks (CNNs) have revolutionized object detection with remarkable accuracy \cite{krizhevsky2012imagenet}. However, the success of these models heavily relies on large annotated datasets for training, which are often costly and time-consuming to acquire. The data scarcity problem poses a significant challenge to the development of robust object detectors that can generalize well to new, unseen objects and domains \cite{tian2019fcos}. 

To address the limitations of data scarcity, considerable research has been devoted to exploring few-shot and zero-shot learning techniques in the field of object detection \cite{alayrac2022flamingo}. Few-shot learning (FSL), in particular, seeks to recognize novel object categories with only a few training examples per class, typically ranging from 1 to 5 \cite{sung2018learning}. The aim is to minimize the prohibitive annotation effort and enable the scalable deployment of object detectors in real-world applications \cite{ravi2016optimization}. By leveraging knowledge transfer and efficient adaptation, FSL methods strive to extract transferable knowledge from a set of base classes with abundant labeled data, enabling generalization to novel classes with limited available examples \cite{wang2020generalizing,song2023comprehensive,xian2018zero}.

Effective FSL algorithms introduce strong inductive biases into models, allowing for rapid adaptation using the limited annotations associated with novel classes. Meta-learning algorithms \cite{nichol2018first}, which train models to quickly adapt to new tasks and metrics with few examples, have shown promise in this regard \cite{li2017meta}. Transfer learning from related domains and data augmentation techniques are also commonly employed to enhance FSL performance \cite{sun2019meta,yu2020transmatch}. Additionally, distance metric learning is utilized to learn embeddings that reflect semantic class relationships, aiding in effective few-shot object detection \cite{jiang2020multi}.

While few-shot classification has been extensively explored, few-shot object detection presents unique challenges \cite{qiao2021defrcn}. In addition to recognizing object classes with limited data, few-shot object detection requires accurate object localization. This localization task becomes particularly challenging when only a small number of examples are available \cite{sun2021fsce}. By overcoming these challenges, FSL techniques have the potential to revolutionize the field of object detection \cite{han2022few}. They can enable accurate and efficient detection of novel objects with minimal annotated data, enhancing the scalability and real-world applicability of object detectors. In this survey, we comprehensively investigate recent advancements in FSL techniques applied to video and 3D object detection, examining their strengths, limitations, and potential for future development.

\subsection{Motivation}
The field of object detection has witnessed significant advancements with the rise of deep learning and convolutional neural networks (CNNs). However, these advancements primarily focus on 2D image-based object detection, which poses limitations in real-world scenarios where objects exist in three-dimensional space and exhibit temporal dynamics \cite{itransition,mindtitan}. Hence, there is a pressing need to explore and understand the progress made in video and 3D object detection. However, existing surveys on FSL have not focused specifically on video or 3D object detection \cite{song2023comprehensive,antonelli2022few,wang2020generalizing,jiaxu2021comparative,li2021concise}.

Video object detection is of paramount importance in various domains such as surveillance, autonomous driving, and action recognition. However, the task of detecting objects in videos presents unique challenges compared to static image-based detection. These challenges arise from the need to cope with motion blur, occlusions, and object interactions across frames \cite{kilitech,mindtitan,itransition,yu2022few}. By conducting a survey specifically dedicated to video object detection, we aim to provide a comprehensive overview of the latest methodologies, techniques, and benchmarks, thus shedding light on the progress made in this critical area and identifying potential future research directions.

On the other hand, 3D object detection, especially in the context of autonomous driving, is crucial for enabling safe and reliable perception systems. Traditional object detection methods primarily rely on 2D sensors such as cameras, which may not provide accurate depth information and struggle with challenging lighting and weather conditions. Integrating LiDAR (Light Detection and Ranging) sensors with cameras can significantly enhance the detection accuracy by providing precise depth information. However, 3D object detection remains a challenging task due to the sparsity of LiDAR point clouds, object occlusions, and the need to handle large-scale 3D data \cite{zhou2018voxelnet,qi2017pointnet++}. Our survey on 3D object detection aims to provide an in-depth analysis of the state-of-the-art techniques, highlighting their strengths, limitations, and novel approaches that address these challenges.

By conducting a survey on both video and 3D object detection, we aim to bridge the gap and provide a comprehensive understanding of the advancements and challenges in these emerging areas. By exploring the latest techniques, model architectures, and evaluation benchmarks, we can assess the progress made, identify gaps in current approaches, and propose potential research directions for future work. This survey serves as a valuable resource for researchers, practitioners, and developers working on video and 3D object detection, paving the way for further advancements in these domains.

\subsection{Organization of the Paper}
\begin{figure*}[htpb!]
    \centering
    \includegraphics[scale=0.2]{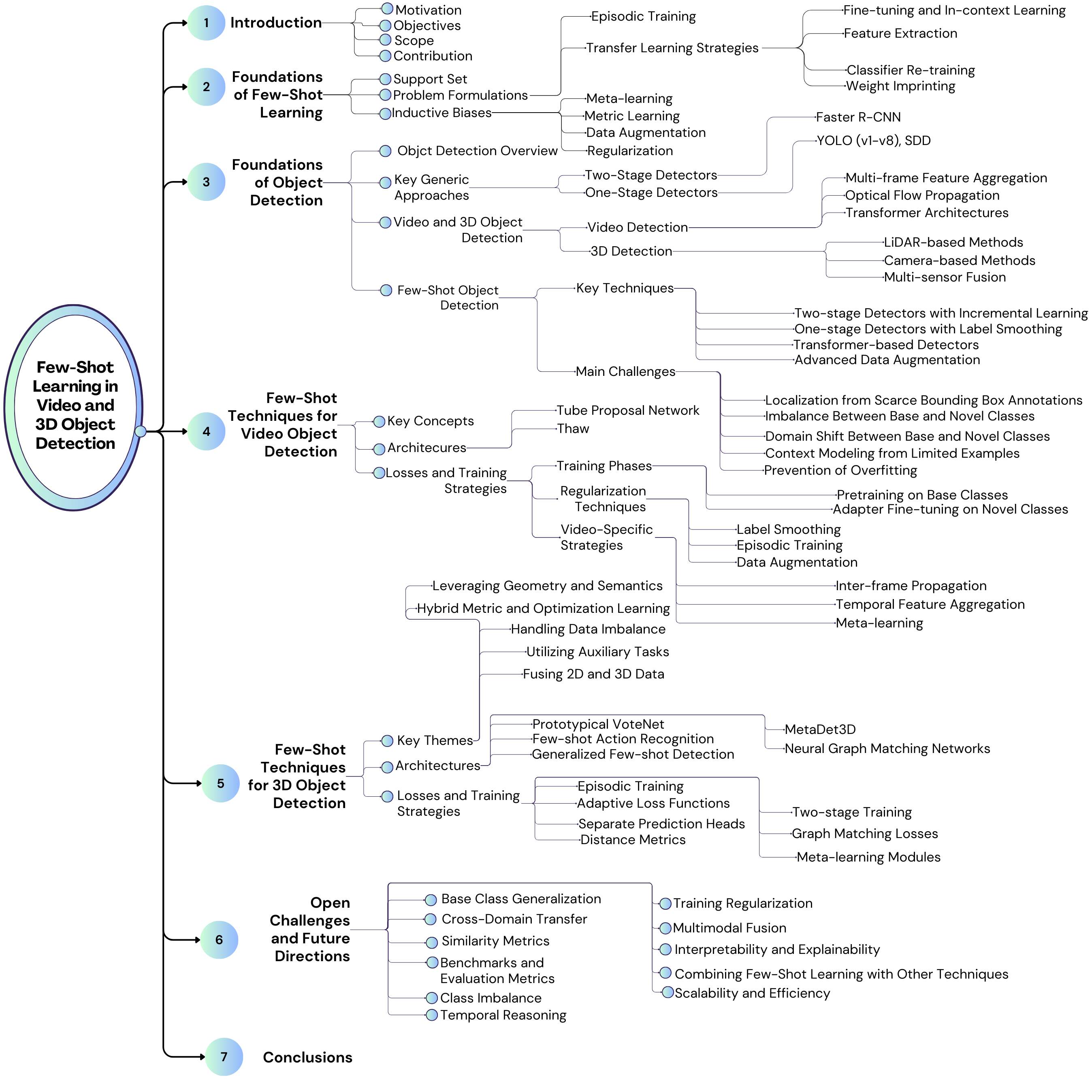}
    \caption{Visual taxonomy illustrating comprehensive structural organization of survey content}
    \label{fig:enter-label}
\end{figure*}

This paper is organized into seven sections as follows: Section 1 provides an introduction that motivates the need for a comprehensive survey on FSL techniques for video and 3D object detection. It highlights the unique challenges posed by these domains and outlines the structure of the paper. Section 2 establishes the theoretical foundations of few-shot learning by reviewing key concepts, problem formulations, and common strategies. It focuses on principles like the support set, episodic training, meta-learning, metric-based approaches, data augmentation, and regularization. Section 3 explores the fundamentals of object detection, including two-stage and one-stage detector paradigms. It analyzes influential architectures like Faster R-CNN, YOLO, and SSD, and examines video and 3D detection approaches. Section 4 provides an in-depth analysis of state-of-the-art few-shot techniques tailored for video object detection. It discusses specialized model architectures, losses, and training methodologies to overcome video-specific challenges. Section 5 investigates few-shot learning strategies for 3D object detection using modalities like LiDAR. It reviews model designs, losses, and training procedures enabling effective few-shot detection on sparse 3D data. Section 6 identifies open challenges and promising research directions to advance few-shot video and 3D object detection. It proposes solutions to limitations in existing approaches. Section 7 presents concluding remarks and summarizes the key insights. Additional architectural diagrams, detailed comparisons, and secondary discussions are provided in the supplementary material. To provide an overview of the paper structure, a visual taxonomy outlining the relationships between the key sections and topics is presented in Figure \ref{fig:enter-label}. This diagram aims to enhance comprehension of the survey scope and content flow for the reader.

\section{Foundations of Few-Shot Learning}
Few-shot learning (FSL) has emerged as a critical research area in deep learning to address the pressing need for vast labeled data, which is often expensive, time-consuming, or infeasible to obtain in real-world scenarios. As deep learning model complexity grows with millions or billions of parameters, substantial data is required to avoid overfitting and ensure generalizability \cite{jiang2022back}. FSL counters this limitation by recognizing new visual concepts from only a few labeled examples, typically 1-5 shots. FSL problems are commonly formulated as classification tasks, where models are provided with scarce labeled examples of new classes called support sets, and must predict labels of unseen query samples from those classes \cite{snell2017prototypical}. Meta-learning algorithms are widely utilized to train FSL models by learning to swiftly adapt to new tasks through experience gained from prior tasks \cite{finn2017model}. Metric-based approaches have also proven effective by learning distance metrics to measure support and query sample similarities \cite{koch2015siamese}. Additionally, transfer learning by pre-training on labeled data can enhance FSL performance \cite{pan2009survey}. This section summarizes the core principles and techniques underpinning FSL, with further details in the \textbf{supplementary document}.

Central to FSL is the sparse support set representing each new class that models must generalize from \cite{boudiaf2020information,wang2020generalizing,zhang2019variational}. Key training strategies include episodic training on simulated few-shot tasks \cite{qiao2019transductive,hajimiri2023strong} and transfer learning to utilize knowledge from data-rich base classes when adapting to novel classes \cite{gavves2015active}. Fine-tuning and in-context learning show promise, but require careful experimental design and tuning \cite{mosbach2023few,eustratiadis2023neural,peng2020fine,liu2022few,shen2021partial,hu2022pushing}. Feature extraction directly applies pre-trained models to novel classes \cite{guo2019spottune,li2023attention,xu2023complementary,xu2023complementary}. Classifier retraining uses base model features to train new classifiers from scratch \cite{shafahi2019adversarially,real2020automl}. Weight imprinting provides informed initialization of novel class weights \cite{abuduweili2021adaptive,yan2023few}. Overall, transfer learning enables utilizing prior base class knowledge when adapting to limited novel data. Vital techniques involve meta-learning algorithms that optimize for rapid adaptation \cite{finn2017model,hospedales2021meta}, metric-based approaches for classification by learned sample similarities \cite{snell2017prototypical,vinyals2016matching,sung2018learning}, data augmentation for regularization \cite{antoniou2017data,shorten2019survey,lemley2017smart}, and explicit regularization to prevent overfitting \cite{goodfellow2016regularization}. Together, these mechanisms provide models with essential generalization capabilities and inductive biases tailored for effective learning under limited supervision, enabling rapid adaptation and knowledge transfer when data is scarce \cite{srivastava2014dropout,kukavcka2017regularization,li2021few}. The \textbf{supplementary document} provides additional details on the support set, problem formulations, training strategies, and inductive biases discussed in this foundations of few-shot learning section.

\section{Foundations of Object Detection}
This section provides an overview of fundamental concepts in object detection, before discussing techniques for video and 3D detection. Further architectural and methodological details are in the \textbf{supplementary document}. 

Object detection integrates classification and localization to identify object categories within images and enclose them in bounding boxes \cite{girshick2014rich}. Given variability in object quantities, initial detection strategies leveraged sliding windows \cite{cheng2023towards}. However, convolutional neural networks (CNNs) now dominate \cite{girshick2015fast,ren2015faster}. Object detection  involves training from supervised datasets containing images $X$ and corresponding annotations $y$ to extract feature maps $F$ that enable bounding box regression and classification. There are two paradigms in object detection: the two-stage detector and the single-stage detector. 

The Faster R-CNN architecture stands out in the two-stage detector category. It integrates ntegrates a Region Proposal Network (RPN) and a Fast R-CNN detection network. The RPN uses a convolutional network to generate object proposals with scores based on a set of anchor boxes. Proposals are reshaped via RoI pooling into fixed features for the detection network. The Fast R-CNN extracts features from the proposals to classify objects and refine bounding boxes. It uses a multi-task loss for classification and regression and separate bounding box regressors per class. 

One-stage detectors directly output object locations and classes in one pass, allowing faster inference but reduced accuracy. YOLO is a seminal one-stage object detector using a single CNN to concurrently predict class probabilities and bounding boxes \cite{redmon2016you}. YOLO divides the input image into an \(S \times S\) grid. Each grid cell predicts $B$ boxes with objectness confidence scores. Predictions comprise box center coordinates, dimensions, and class. While exploiting contextual information, YOLO's grid approach can miss small objects. YOLOv2 introduced anchor boxes and multi-scale training to address this \cite{redmon2017yolo9000}. YOLOv3 incorporated a deeper architecture and multi-scale predictions to boost accuracy while maintaining real-time performance  \cite{redmon2018yolov3}. YOLOv4 optimized speed and accuracy via techniques like weighted residual connections, cross-stage connections, normalization, and self-adversarial training \cite{bochkovskiy2020yolov4}. Open-source YOLOv5 refined efficiency and usability \cite{yolo5}. YOLOv6 adopted an anchor-free design optimized for industrial use cases \cite{li2022yolov6}. YOLOv7 pushed accuracy further, surpassing prior detectors across FPS targets without pre-trained backbones, via innovations in self-supervised learning, model design, and enhancements \cite{wang2023yolov7}. Most recently, YOLOv8 introduced anchor-free prediction with fewer boxes and faster NMS, achieving state-of-the-art accuracy by disabling aggressive augmentation during late training \cite{yoloV8}. 

SSD enhances YOLO by utilizing anchor boxes tailored to diverse object shapes and performing detection across multiple network layers to achieve robustness across varying scales  \cite{liu2016ssd}. Smaller feature maps in earlier layers focus on detecting larger objects, while higher resolution layers target smaller objects. This multi-scale design contrasts with YOLO's single output scale, enabling SSD to capture a wide range of object sizes. Predictions from all layers are aggregated and refined to produce unified detections across scales. SSD's multi-feature map architecture has influenced other single-stage detectors for handling scale variation through its effectiveness at detecting objects across a spectrum of sizes. 

\subsection{Video and 3D Object Detection}
Video object detection involves identifying and localizing objects across consecutive frames in video sequences, presenting distinct challenges compared to static image detection, including motion blur, defocus, complex object motions, and viewpoint variations over time \cite{he2020temporal}. Effective techniques require specialized modeling of temporal information propagation and consistent detection across frames to cope with video-specific complexities \cite{lin2021multi}. Key techniques for video object detection harness temporal context to enhance per-frame detection accuracy. One approach is temporal aggregation which propagates detections across frames using optical flow \cite{cores2023spatiotemporal,zhu2017flow} or aligns and averages neighboring frame features \cite{sun2022coarse,xu2022multilevel}. This provides useful contextual clues to help resolve detection ambiguities \cite{sun2022coarse,honari2016recombinator}. Spatial aggregation is another strategy which applies larger receptive fields or coarse pooling to frames farther from the reference frame, organizing multi-scale features and improving inter-frame complementarity. Flow-guided aggregation employs optical flow correspondence \cite{guo2021frame} to enable flexible multi-frame fusion at earlier layers before final detection \cite{muralidhara2022attention,han2020exploiting}, although computational costs and handling large motions remain challenges \cite{guo2021frame,qian2020adaptive}. Recently, transformer-based architectures like TransVOD \cite{zhou2022transvod} and DETR \cite{carion2020end} have shown promising results by enabling effective modeling of long-range dependencies through self-attention, achieving state-of-the-art accuracy \cite{wang2021end,he2021end,liu2021swin,cui2023feature}. Combining transformers and multi-frame feature aggregation is also being explored to jointly leverage temporal context, inter-frame correlations, and self-attention \cite{cui2023faq,muralidhara2022attention}. However, balancing efficiency remains an open issue.

Specialized techniques have emerged to address the unique complexities of 3D object detection. LiDAR-based 3D detection operates on raw point clouds using networks like PointNet \cite{gao2023pointpainting}, extended by works like PointRCNN \cite{shi2019pointrcnn}, Part-A2 Net \cite{shi2019part}, and PV-RCNN \cite{shreyas20213d} for proposal generation and refinement. Other approaches aggregate points into efficient pillars, like PointPillars \cite{lang2019pointpillars} and PIXOR \cite{tian2022fully}, but lose details. Advanced pillar variants like SpindleNet \cite{zhao2017spindle} and CenterPoint \cite{yin2021center} improve representations by encoding local context more effectively. Camera-based 3D object detection such as 3DOP \cite{kim20203d} lifts 2D detections into 3D or estimates depth to apply LiDAR techniques, enhanced by stereo fusion as observed in Mono3D \cite{chen2016monocular}, Mono3D++ \cite{he2019mono3d++}, and Pseudo-LiDAR \cite{you2019pseudo}. Earlier works rely heavily on priors and ground plane assumptions \cite{liu2023real,ku2018joint}. Multi-sensor fusion combines LiDAR geometry and camera semantics, with robust recent approaches exploring transformer soft-attention. Early fusion integrates modalities in the network while late fusion generates proposals in one modality using the other. PointPainting \cite{vora2020pointpainting} and PointFusion \cite{xu2018pointfusion} feature learned feature fusion throughout. Key challenges include computational efficiency, maintaining geometric details, handling noise and occlusion, optimizing proposal generation, and effectively fusing multi-modal cues. Specialized techniques aim to address the unique complexities of sparse 3D data for robust detection vital for applications like autonomous driving. Further architectural and methodological details are provided in the \textbf{supplementary document}.

\subsection{Few-Shot Object Detection}
Few-shot object detection poses unique challenges compared to classification, demanding accurate localization from extremely scarce bounding box annotations. Popular techniques like incremental learning in two-stage detectors \cite{belouadah2023multiod}, label smoothing in one-stage detectors \cite{krothapalli2020one,lv2023detrs}, specialized augmentation \cite{su2022dsla,liu2023recent}, and transformer architectures \cite{jin2023incremental} help address these difficulties. However, core issues persist, including unreliable localization \cite{jiang2023few,qiao2021defrcn}, imbalance between base and novel classes \cite{kohler2023few,wu2021generalized}, complex domain shifts \cite{jiaxu2021comparative,kang2019few,shangguan2023identification}, lack of context from limited examples \cite{liu2023recent,wang2020frustratingly,wang2022few}, and overfitting tendencies  \cite{kohler2023few,liu2023recent}. Advanced data augmentation shows promise but faces information-theoretic constraints on synthesizing new signals from scarce data. Further innovations in areas like meta-learning, metric-based learning, context modeling, and transfer learning hold promise for advancing few-shot detection by overcoming limitations like scarce annotations, class imbalance, and domain shifts. The \textbf{supplementary document} provides additional details on few-shot object detection challenges, and state-of-the-art techniques.

Building on these foundations, the following sections dive deeper into applying FSL to address the unique complexities posed by video and 3D object detection.

\section{Few-Shot Video Object Detection}
In the context of video, FSL becomes especially valuable given the additional difficulty of annotating objects across multiple frames. Manually labeling objects across numerous video frames is much more laborious than for static images. Few-shot video detection techniques can help significantly reduce annotation requirements by propagating information across frames.

\subsection{Key Strategies in Few-Shot Video Object Detection}
Few-shot video object detection (FSVOD) task presents unique challenges compared to few-shot image detection, as it requires effectively modeling complex spatiotemporal variations in object appearance, scale, motion, and viewpoint across frames. To address these challenges, several key strategies have been developed in the field of FSVOD.

One common approach in FSVOD is to use a pretrained CNN backbone to extract rich spatiotemporal features from input video clips. This backbone network captures both spatial information and temporal dynamics, enabling the model to analyze object appearance and motion across frames. Additionally, modeling local object cues and global context throughout the video is important. Specialized components, such as memory modules, region proposal networks, and temporal propagation mechanisms, have been developed to enhance coherence across frames by reasoning about object trajectories and identities. Metric-based learning approaches are also commonly used in FSVOD. These approaches compare query and support embeddings to measure the similarity between objects in the video. By leveraging metric learning, FSVOD models can effectively match objects and adapt to novel classes from limited support examples. Furthermore, meta-learning focused fine-tuning strategies have been developed to enable rapid adaptation to novel classes while retaining knowledge from the base classes. This allows the FSV to quickly learn and generalize from few-shot examples. Techniques like multi-scale feature learning, relational reasoning, metric-based matching, and meta-adaptability enable FSVOD models to recognize novel object classes from scarce video examples by effectively capturing spatiotemporal information and adapting to novel classes with limited support examples.

\subsection{Architectures for FSVOD}
Recent few-shot video object detection architectures follow a two-stage approach. In the first stage, a proposal generation module creates spatiotemporal tube proposals representing object trajectories across frames. The second stage classifies these aggregated tube features by matching against the few-shot support examples to produce detection predictions. Two representative two-stage architectures are the Tube Proposal Network (TPN) \cite{fan2022few} and the MEGA model-based Thaw \cite{yu2022few}, which effectively incorporate this proposal-classification framework tailored for few-shot video scenarios. The TPN architecture generates tube proposals connecting objects across frames to utilize temporal consistency. Thaw's two-stage design aggregates both local object features and global video-level features to classify tube proposals based on comparison with the few-shot supports. These concrete implementations showcase how the two-stage approach of proposal generation followed by temporal feature aggregation and matching enables state-of-the-art few-shot detection performance on videos.

\subsubsection{TPN and TMN + Hybrid}
Fan et al. \cite{fan2022few} proposed the tube proposal network (TPN) as a representative architecture for few-shot video object detection. As illustrated in Figure S1 of the supplementary document, their system comprises various components for both training and inference, including the Tube Proposal Network (TPN), Temporal Alignment Branch (TAB), and Query and Support branches. During training, the weight-shared convolutional neural network (CNN) backbone extracts spatiotemporal features from the input query video frames and support images. The backbone enables subsequent analysis and detection by learning discriminative representations. 

Specifically, the query branch processes two query frames \( \{I_1, I_2\} \) using RoIAlign to extract query features \( \{Q_{i1}, Q_{i2}\} \) corresponding to each proposal \( p_i \) of instance \( i \). These instance-specific query features capture relevant visual cues about the object. Concurrently, the support branch extracts support features \( S \) from the ground-truth boxes in the support images to serve as references during matching. Crucially, the Temporal Alignment Branch (TAB) aligns the query features temporally to ensure synchronization across frames before comparison with the support features. Matching occurs in the tube matching network (TMN), which utilizes tube-level features aggregated over time by the TPN via inter-frame regression and identification scoring. This establishes temporal consistency in detections. The matching network, called the Multi-Relation Network, compares the aggregated query features \( Q = \frac{Q_{i1} + Q_{i2}}{2} \) and support features \( S \) by computing their distance, measuring similarity between them. Additionally, a multi-relation head and contrastive training, inspired by FSOD, improve the discriminability of the matching. This allows effective classification of the query features based on their affinity to the supports. By integrating and jointly optimizing the TPN and TMN end-to-end, the model can handle the challenges of high dynamism and diversity in video object detection.

\subsubsection{Thaw}
A different architecture was created by \cite{yu2022few} for few-shot video object detection, as shown in Figure S2 of the supplementary document. This framework is called Thaw and consists of several essential steps:

\begin{enumerate}
    \item \textbf{Pretraining Phase:} The first stage of the proposed method involves pretraining a video object detector on a base dataset containing an abundant number of videos per class \( V_{\text{base}} = \{V^{i}_{\text{base}} | i = 1,...,N\} \), where \( N \) is the number of videos. The MEGA model \cite{chen2020memory} is utilized as the video object detector because it can efficiently aggregate both local and global spatiotemporal information across frames in a video.

Specifically, for each key frame $I_k$ in a given video, MEGA generates multiple feature representations. First, a local feature pool $L$ is extracted from region proposals in $I_k$. Next, a global feature pool $G$ is obtained by applying a convolutional network backbone on the entire frame $I_k$. An aggregated local feature pool $L_g$ is then formed which condenses information from $L$ across multiple neighboring key frames. Additionally, an enhanced local feature pool $L_m$ is generated by integrating $L$ and $G$. Finally, a memory module $M$ aligns features from region proposals using ROI alignment.

These various features are concatenated into an enhanced feature representation $f_e(I_k)$ for each key frame $I_k$:

$$f_e(I_k) = {f_e(I_k)_1, f_e(I_k)_2,...,f_e(I_k)_Q}$$

where $Q$ is the dimensionality of the concatenated feature vector. This enhanced feature representation $f_e(I_k)$ is then utilized in MEGA's region proposal network for object classification and localization in the video. The multi-level feature extraction provides both local object details and global spatiotemporal video context to enable effective few-shot detection.

\item \textbf{Adaptation Phase:}  Subsequent to the pretraining, the model is adapted to novel classes using a few-shot dataset with limited videos per class \(V_{\text{novel}}\). A cosine classifier is introduced in the detection head:
\[
S(W,x) = \left[ \cos(\theta(w_1, x)), \dots, \cos(\theta(w_{N+M}, x)) \right]'  
\]
where \(W\) contains class weight vectors \(w_i\) and \(S(W,x)\) measures similarity between features \(x := f_e(I_t)_l\) (where \(f_e\) is the feature extractor and \(I_t\) is the input frame at time \(t\)) and classes. The probability for the \(i^\text{th}\) class can then be calculated as: 
\[
P_i = \frac{\exp(S(W,x)_i)}{\sum_c \exp(\alpha S(W,x)_c)}
\]
where \(\alpha\) is a scaling factor to reduce the discrepancy between one-hot and real distributions [1], [20], [30].

\item \textbf{Fine-tuning Phase:} In the final phase, fine-tuning uses Joint (all weights updated), Freeze (only classifier updated), and Thaw (gradual unfreezing) methods:
\[
\text{Freeze } f_e(\cdot) \rightarrow \text{Unfreeze } f_e(\cdot)
\]
The Joint method fine-tunes all weights, but often leads to overfitting on small datasets. Freeze only updates the classifier while keeping the feature extractor fixed, making it suitable for FSL. Thaw gradually unfreezes the feature extractor for improved adaptation. Recent work shows Freeze attains the highest few-shot detection performance to date by preventing overfitting to the limited novel class examples during fine-tuning. While fully fine-tuning tends to overfit on scarce data, Freeze provides a simple yet effective alternative that concentrates model updates only on the task-specific classifier head during few-shot adaptation \cite{lee2022rethinking}.

\end{enumerate}

Additionally, a balanced sampling strategy is proposed to overcome the class imbalance between novel and base classes during fine-tuning. Classes are uniformly sampled during each iteration to provide an evenly distributed gradient update. This prevents the model from overfitting to base classes and forgetting novel classes. Experiments show balanced sampling is crucial for good few-shot detection performance.

\subsection{Losses and Training Strategies for Few-Shot Video Object Detection}
Achieving effective FSL on videos requires a specialized training methodology that accounts for the unique spatiotemporal dynamics of these data. The following section discusses key training phases, regularization techniques, and video-specific training strategies that contribute to accurate and generalizable few-shot detection with limited novel class training data.

\subsubsection{Training Phases}
A multi-phased training approach is critical to prevent overfitting generalization in few-shot video object detection. The training phases typically include:

\begin{itemize}
    \item \textbf{Pretraining Phase:} In this phase, the feature extractors are pretrained exclusively on base classes to learn transferable representations. By leveraging abundant labeled data from the base classes, the models can extract high-level features that capture relevant visual patterns and semantics. This base class knowledge transfers well when adapting to novel classes, allowing the models to generalize effectively with limited labeled data.
    \item \textbf{Adapter Fine-Tuning Phase:} The task-specific components, such as matching networks or tube proposal modules, are fine-tuned on the novel classes while keeping the base class weights fixed. This approach prevents interference between the base and novel classes, as the models continue to rely on the prelearned feature representations from the pretrained feature extractors. Gradual unfreezing of later layers in the feature extractors can strike a balance between retaining generalization and increasing model capacity for the novel classes.
\end{itemize}
The multi-phased training approach allows the models to effectively utilize the knowledge acquired from the base classes while adapting to the few-shot novel classes. It helps prevent overfitting and ensures that the models can generalize well to unseen classes in the video data.

\subsubsection{Regularization Techniques}
To further reduce overfitting during the training of few-shot video object detection models, various regularization techniques can be employed.

\paragraph{Label Smoothing} 
Label smoothing is a highly effective regularization technique that can significantly enhance few-shot video object detection models. By introducing small amounts of target noise, label smoothing serves to prevent models from making overconfident predictions solely based on limited video training examples, thereby improving calibration and generalizability \cite{muller2019does}. Recent research has actively explored the use of label smoothing in the context of few-shot video detection tasks. In the case of FSL, where only scarce labeled examples are available for novel classes, models often encounter challenges with overfitting and struggling to robustly detect new classes \cite{han2021query, su2022dsla}. Label smoothing plays a crucial role in mitigating these issues by redistributing some target probability to non-ground truth classes. This reduces the model's reliance on memorization and fosters a more comprehensive and adaptable understanding of the data. The importance of label smoothing further amplifies in the context of few-shot video detection, where each class possesses a limited number of annotated frames, and objects may exhibit significant appearance variations across frames and viewing angles. By discouraging overconfidence, label smoothing compels models to place more emphasis on invariant class-specific features instead of relying on superficial cues. Additionally, label smoothing aids in addressing imbalanced classes \cite{han2021query} commonly observed in few-shot video detection. Given that novel classes typically have far fewer examples than base classes, smoothing techniques effectively limit the model's reliance on individual samples, which prevents biases and enables more balanced and generalizable recognition across both base and novel classes.

\paragraph{Episodic Training} 
Episodic training is a essential technique for improving few-shot performance in video object detection. By constructing varied few-shot task distributions, episodic training exposes the model to diverse training scenarios, enabling better generalization. This training approach organizes the model training into a series of learning problems or episodes, with each episode mimicking the FSL setting encountered during evaluation. Each episode consists of a small training set and a validation set. The model is trained on these small but varied episodes, allowing it to improve its ability to generalize to new tasks with only a few examples during testing. In the context of few-shot video object detection, episodic training has shown promise by constructing episodes that contain only a few labeled frames per video. The model is trained to detect objects in these sparse labeled videos, effectively leveraging information across frames and learning to generalize from limited annotation. Compared to fully supervised pre-training, episodic training better simulates the intended few-shot test scenario. Although originally proposed for image classification, episodic training has proven effective in improving generalization for few-shot video recognition \cite{yu2022few}. By exposing the model to varied few-shot episodes during training, episodic learning encourages the development of inductive biases tailored for rapid adaptation from scarce video data. Overall, constructing representative episodes is a vital technique for enhancing few-shot performance in video understanding tasks.

\paragraph{Data Augmentation} Data augmentation plays a vital role in enabling few-shot video object detection models to generalize effectively from limited labeled data. While basic augmentations such as random cropping, padding, flipping, and color transforms are commonly used \cite{niklaus2017video,xu2023comprehensive}, more advanced techniques like mixup offer the opportunity to combine samples from different classes, thereby exposing models to a more diverse range of augmented samples during training. Dynamic Video Mixup \cite{wu2022dynamic}, for instance, fuses videos from different domains to enhance cross-domain generalization, while Manifold Mixup \cite{mangla2020charting} creates mixes that are robust to small shifts in the data distribution. Additionally, Hard Mixup \cite{roy2022felmi} utilizes uncertainty measures to generate challenging class combinations. These mixup approaches contribute to increased diversity and improved generalization capabilities. Furthermore, beyond mixup, additional advanced augmentation techniques have proven to be effective. For instance, CutMix \cite{nakamura2022few} blends object patches between videos to introduce variations in context, while CutBlur \cite{yoo2020rethinking} incorporates Gaussian blurring to simulate motion and occlusion. Temporal crop and paste \cite{zhang2022unsupervised} perturbs object motion and timing by cropping object tubes and inserting them at different temporal locations in the video. Temporal jittering alters frame rates, improving robustness to variable frame rates during inference. Spatial jittering applies transformations such as translation, flipping, rotation, and scaling to individual video frames, bolstering robustness to spatial variations. Video mixup combines full clips from different domains, which is especially useful for cross-domain few-shot detection \cite{aich2023cross}. Finally, context augmentation involves pasting detected objects from the same classes into new background scenes and contexts, enhancing context invariance for the model \cite{olsson2021classmix}.

\subsubsection{Video-Specific Training Techniques}
To enable more effective FSL in video object detection, several strategies leverage the unique spatiotemporal characteristics of video data. These techniques aim to improve detection consistency, reduce noise, and exploit temporal context.  

\paragraph{Inter-Frame Propagation} In the context of few-shot video object detection, inter-frame propagation is a technique that enhances detection consistency and incorporates valuable temporal context by propagating object detections or features across frames. Recent works have proposed several advancements in inter-frame propagation techniques. Chakravarthy et al. \cite{chakravarthy2021object} proposed a method that utilizes inter-frame attentions for temporally stable video instance segmentation. By refocusing on missing objects using box predictions from neighboring frames, their method overcomes missing detections and improves temporal stability. In another work, Xu et al. \cite{liu2022temporal} proposed a method called Temporal Consistency learning Network (TCNet) for video super-resolution that employs fine-tuned flow estimation and temporal self-alignment modules for motion compensation, demonstrating the effectiveness of inter-frame propagation. Wang et al. \cite{deng2023identity} introduced the Dynamic Warping Network (DWNet) that adaptively warps inter-frame features to improve semantic video segmentation performance, further evidencing the utility of propagation. Zhang et al. \cite{zhang2022optical} combined weighted optical flow prediction with an attention model for object tracking, showing inter-frame propagation's usefulness in tracking. Finally, Lin et al. \cite{lin2022unsupervised} proposed an unsupervised flow-aligned sequence-to-sequence learning approach for video restoration using optical flow for motion compensation. Together, these advancements demonstrate inter-frame propagation's effectiveness for various video tasks like few-shot detection, instance segmentation, super-resolution, semantic segmentation, tracking, and restoration. By propagating information between frames, consistency, context, and performance can be enhanced despite limited supervision.

\paragraph{Temporal Feature Aggregation} Aggregating features over tubes or temporal segments allows models to capture rich contextual information and exploit temporal dynamics of objects. Strategies like temporal average/max pooling \cite{du2021revisiting}, LSTMs \cite{ma2023ms}, and attention mechanisms \cite{jeune2022unified,pal2023temporal,gordevivcius2009parsimonious} provide aggregation across clips or tubelets. These enable models to learn robust spatiotemporal representations \cite{fu2020spatiotemporal,dai2022spatio,jin2021spatiotemporal}, facilitating few-shot detection without requiring extensive annotation \cite{jeune2022unified}. Architectures can be designed to enable aggregation at multiple levels \cite{ma2023ms,he2022temporal}, from early convolutional features to late detection features \cite{nirthika2022pooling,luo2020expectation}. Average pooling reduces the effect of noisy features but can lose prominent features like edges. In contrast, max pooling extracts pronounced features but may overfit more easily \cite{nirthika2022pooling}. To balance these tradeoffs, mixed pooling combines max and average pooling. More advanced pooling explores higher order statistics like skewness and kurtosis \cite{rouvier2021study}. Tree pooling and stochastic pooling add randomness to avoid overfitting \cite{zafar2022comparison}. Spatial pyramid pooling adapts pooling to spatial structure. Another notable approach is the use of spatiotemporal graph networks, which make use of graph convolutions and recurrent neural networks (RNNs) to incorporate both spatial and temporal information \cite{jin2021spatiotemporal}.

\paragraph{Meta-learning} Recent advancements in few-shot video detection have focused on leveraging meta-learning to enable models to quickly adapt to novel classes with only a few examples. Specifically, meta-learning can take advantage of the additional spatial-temporal information present in videos compared to static images \cite{vu2022few,yu2022few}. One approach is to pretrain a video object detector on a base dataset by aggregating local and global information across frames using techniques like MEGA \cite{yu2022few}, and then fine-tune it on the novel classes. The model learns to effectively extract spatial-temporal features from the base classes that transfer well to novel classes. Another promising direction is to integrate spatial reasoning into the few-shot video detection framework \cite{kim2021spatial}. For example, STEm-Seg \cite{kim2021spatial} encodes relative spatial contexts between tubelet proposals in a graph neural network. This allows the model to understand object interactions and scene layout to generalize better. In addition, recent work has explored going beyond individual frames to use information from surrounding frames when adapting to novel classes \cite{vu2022few,yu2022few}. For example, TPN \cite{yu2022few} aggregates RoI features from a local temporal window centered on each query frame during few-shot matching. This provides useful cues from motion and temporal consistency to recognize novel objects with scarce examples. However, there remain significant challenges in scaling up to longer videos and more complex scenes. Further advancements in meta-learning will help enable few-shot video detection for real-world applications.

Various specialized loss functions and training strategies have been developed to enable effective few-shot learning on videos. To provide readers with an overview of these techniques, we include the following comparison table \ref{table:extended_comparison} summarizing the key methods discussed in this survey. This table highlights how contemporary approaches tailor their optimization methodology to account for challenges like class imbalance and limited supervision.

\begin{table*}[htb!]
\centering
\caption{Comparison of loss functions, training strategies, and class imbalance techniques for few-shot video object detection (including generic video object detection techniques)}
\label{table:extended_comparison}
\begin{tabularx}{\textwidth}{lXXXX}
\toprule
\textbf{Method} & \textbf{Loss Function} & \textbf{Aux. Losses} & \textbf{Training Strategy} & \textbf{Class Imbalance Tech.} \\
\midrule
TPN [30] & Cross-entropy & Segmentation loss & Episodic training & - \\
\addlinespace
Thaw [178] & Cosine similarity & - & Balanced sampling & Balanced sampling \\
\addlinespace
FSCE [141] & Online hard mining & - & Adam optimizer & Focuses on hard examples \\
\addlinespace
TCL [27] & Consistency loss & - & Information propagation & Inter-frame propagation \\
\addlinespace
DSLA [140] & Smooth L1 & - & SGD optimizer & Label smoothing \\
\addlinespace
FSOD [156] & Online hard mining & Attribute prediction loss & Class re-weighting & Feature re-weighting \\
\addlinespace
MetaYOLO [149] & MSE & - & Meta-learning & Over-sampling \\
\bottomrule
\end{tabularx}
\end{table*}

Table \ref{table:extended_comparison} compares several major few-shot video object detection methods in terms of their loss functions, auxiliary losses, training strategies, and techniques to handle class imbalance. The cross-entropy, cosine similarity, online hard example mining, consistency, and smooth L1 losses are common choices adapted to the few-shot setting. Auxiliary losses like segmentation help improve feature learning. Strategies like episodic training, balanced sampling, label smoothing, and information propagation aim to prevent overfitting and make use of the spatiotemporal structure of videos. Re-weighting and over-sampling help mitigate issues with class imbalance. As shown in Table \ref{table:extended_comparison}, specialized loss formulations and training methodologies are instrumental for achieving effective few-shot learning for video object detection. The multi-faceted approach of combining tailored losses, auxiliary tasks, regularization techniques, and class re-balancing enables models to generalize from scarce training data across imbalanced classes. Advancing these optimization and learning strategies remains an active research area for improving few-shot video object detection.

\section{Few-Shot 3D Object Detection}
Few-shot 3D object detection (FS3DOD) stands at the intersection of 3D computer vision and FSL, aiming to detect objects in 3D space with minimal labeled examples. The challenge is intensified due to the inherent complexities of 3D data, such as point clouds from LiDAR or depth sensors, which are inherently sparse, unordered, and lack the rich texture information available in 2D images.

\subsection{Key Themes in FS3DOD}
One recurring theme across these algorithms is the emphasis on leveraging both geometric and semantic information. Many FS3DOD approaches build upon PointNet-based architectures, which are adept at handling raw point clouds, extracting hierarchical features and preserving the spatial structure of the data. These architectures often employ attention mechanisms, prototype matching, and other techniques to enhance the discriminative power of the learned embeddings.

Furthermore, there's a trend towards hybrid models that synergize both metric-based and optimization-based FSL strategies. For instance, some methods use prototype-based approaches where class representations are computed as the mean of feature embeddings. These prototypes are then used to classify query points based on their similarity, often measured through cosine distances or other distance metrics.

Another significant insight is the challenge of data imbalance in the few-shot setting. Several methods introduce novel loss functions or sampling strategies to handle the disparity between base classes with abundant data and novel classes with limited examples. These strategies aim to prevent the model from being overwhelmingly biased towards the base classes.

Additionally, the role of auxiliary tasks, such as segmentation or attribute prediction, is evident in many FS3DOD algorithms. By training on these auxiliary tasks alongside the primary detection task, models can learn richer and more generalized feature representations.

The fusion of 2D and 3D information is also a promising direction. Some algorithms project 3D point clouds into 2D space, extract features using 2D CNNs, and then lift these features back into 3D space for detection. This multi-modal approach aims to capitalize the strengths of both 2D images and 3D point clouds.

In summary, FS3DOD represents a confluence of techniques designed to address the unique challenges posed by 3D data and the scarcity of labeled examples. As the demand for 3D object detection in applications like autonomous driving, robotics, and augmented reality continues to grow, the innovations in FS3DOD provide a promising pathway to achieve robust performance with minimal annotations.

\subsection{Architectures}
Most FS3DOD build on top of standard 3D convolutional backbones like VoxelNet \cite{zhou2018voxelnet} or PointNet++ \cite{qi2017pointnet++} to extract features from raw point clouds or voxel grids. The extracted features are then fed into metric learning modules for comparison against few-shot prototype features to produce classifications. Some prominent FS3DOD architectures that follow this overall pipeline are described below for better understanding.

\subsubsection{Prototypical VoteNet for FS3DOD}
Prototypical VoteNet is a novel methodology introduced to address the challenges inherent in 3D point cloud object detection \cite{zhao2022prototypical}. Traditional approaches in this domain heavily depend on a vast amount of labeled training data. However, acquiring these labels is both expensive and time-intensive. This is particularly challenging when considering the detection of objects from novel categories, for which only a limited number of labeled examples might be available. To circumvent these challenges, researchers proposed the Prototypical VoteNet \cite{zhao2022prototypical}, which aims to efficiently detect and localize instances even with minimal training data.

The core innovation of Prototypical VoteNet lies in its introduction of two distinct modules namely Prototypical Vote Module (PVM) and Prototypical Head Module (PHM) as shown in Figure S3 of the supplementary document.

\textbf{Prototypical Vote Module (PVM)}
The PVM is designed to take advantage of shared 3D basic geometric structures among object categories. Recognizing that these structures can be class-agnostic, the PVM focuses on refining the local features of novel categories based on these commonalities. It consists of the following key components:
\begin{itemize}  
    \item \textbf{Memory Bank Construction}: A class-agnostic memory bank \( G = \{g_k\}_{k=1}^K \) is constructed, containing geometric prototypes. These prototypes are learned from the rich base categories.
    
    \item \textbf{Prototype Update Mechanism}: Initialized randomly, the prototypes undergo iterative updates during the training process. The formula for this is given by:
    \[ g_k \leftarrow \gamma \cdot g_k + (1 - \gamma)f_k \]
    where \( f_k \) is the average of point features \( \{f_m\}_k \) assigned to prototype \( k \) and \( \gamma \) is a momentum term.
    
    \item \textbf{Feature Refinement}: PVM employs a multi-head cross-attention module to enhance the input point features using the established prototypes. The refinement formula is:
    \[ f_j \leftarrow \text{Cross\_Att}(f_j, \{g_k\}) = \sum_{h=1}^{H} W_h \left( \sum_{k=1}^{K} A_{h,j,k} \cdot V_h g_k \right) \]
    Where \( f_j \) represents the point feature, \( g_k \) signifies the prototype, and \( A_{h,j,k} \) is the attention weight that measures the similarity between the query \( f_j \) and key \( g_k \).
    
    \item \textbf{Vote Layer}: The refined features are subsequently used by the Vote Layer, which predicts point offsets and features.
\end{itemize}

\textbf{Prototypical Head Module (PHM)}
The Prototypical Head Module (PHM) plays a crucial role in few-shot 3D detection by utilizing class-specific prototypes to refine object features. These prototypes, denoted as \(E = \{e_r\}_{r=1}^R\), are extracted from a support set, where $R$ indicates the total number of class prototypes available.

The primary purpose of the PHM is to enhance object features by leveraging class-specific prototypes. To achieve this, the PHM employs a two-step process. First, it extracts the prototype for a specific class \(e_r\) by averaging the instance features from support samples of that class. This class prototype captures the representative characteristics of the objects belonging to that class. Next, the PHM utilizes a multi-head cross-attention module, similar to the Prototypical VoteNet's (PVM) approach, to refine the object features. The refinement is accomplished by applying the cross-attention mechanism as follows:

\[ f_{o,t} \leftarrow \text{Cross\_Att}(f_{o,t}, e_r) \]

In this equation, \(f_{o,t}\) represents the object feature, and \(e_r\) denotes the class prototype. By combining the object feature with the class-specific prototype, the PHM enhances the discriminative power of the object representation.

After feature refinement, the enhanced features are passed to the prediction layer, which is responsible for the actual detection process. The prediction layer utilizes the refined features to make accurate predictions for object presence and location. To train the PHM module, an episodic training approach is employed. This training strategy is designed to learn a distribution of few-shot tasks. By exposing the PHM to various few-shot scenarios during training, it can effectively generalize and adapt to new objects with limited annotated examples.

In summary, the Prototypical Head Module (PHM) in Prototypical VoteNet takes a dual-pronged approach to few-shot 3D detection. While the PVM refines local features through geometric prototypes, the PHM focuses on enhancing global features by utilizing class-specific prototypes. This combination enables the model to effectively handle the challenges of few-shot 3D object detection, such as sparsity and lack of texture, by leveraging both local and global information.

\subsubsection{Generalized Few-Shot 3D Object Detection}
Figure S4 shows the overall framework for generalized few-shot 3D object detection \cite{liu2023generalized}. The input 3D point cloud first goes through a 3D feature extractor based on VoxelNet to generate feature embeddings. These features are then processed by a region proposal network (RPN) for further feature encoding. The features then pass through a shared convolutional layer, whose outputs are fed into multiple prediction heads for final detection.

The framework adopts a two-stage training approach. In the base training stage, it trains on the base classes with abundant data. In the few-shot fine-tuning stage, it freezes the base network and adds incremental branches for novel classes, each with a small training set. Specifically, each incremental branch for a novel class contains a convolution layer, a batch normalization layer, and a ReLU activation layer. These branches share the feature embeddings from the earlier layers, but make separate predictions for their respective classes.

During fine-tuning, the loss function is a weighted combination of a sample adaptive balance (SAB) loss \( L_{\text{SAB}} \) for classification and an \( L_1 \) loss for regression:
\[
L = L_{\text{SAB}} + \lambda L_{\text{regression}}
\]
where \( \lambda \) balances the two loss terms.

The SAB loss handles the imbalance between foreground objects and background regions, and focuses on hard negative samples that have high confidence scores. It dynamically adjusts weights \( w_{\text{pos}}, w_{\text{neg}}, w_{\text{hn}} \) for positive, negative, and hard negative samples respectively based on the number of samples.

Figure S5 shows the incremental branches added for novel classes. Each novel class gets its own branch that shares an earlier convolutional layer with the base class branches. This avoids interference between base and novel classes during fine-tuning. Only the novel class branches are updated during the second training stage.

\subsubsection{MetaDet3D}
MetaDet3D is a meta-learning based framework for few-shot 3D object detection introduced by \cite{yuan2022meta}. It takes a novel approach of using meta-learning to derive class-specific knowledge from the few-shot support examples, which is then used to guide the downstream 3D object detector. Specifically, MetaDet3D comprises two essential components that operate in collaboration, as shown in Figure S6 and described below:
\begin{itemize}
    \item \textbf{3D Meta-Detector:} The first is a lightweight 3D Meta-Detector implemented as a class-specific reweighting module $M$. It takes as input the few support examples available for each novel class. A PointNet++ backbone extracts features from these support points.  The meta-detector then condenses these features into a compact class-specific reweighting vector $z_n$ for each novel class $n$. This reweighting vector encapsulates class-specific knowledge learned from the scarce support examples:

\begin{equation*} z_n = M(\text{support samples}) \end{equation*}

    \item \textbf{3D Object Detector:} The second component is the primary 3D Object Detector, which uses the reweighting vectors to guide its prediction process. This component consists of three sub-components: point feature extraction, guided voting and clustering, and guided object proposal.
    \begin{itemize}
        \item \textbf{Point Feature Extraction:} The PointNet++ backbone $F$ is used to extract point features $[x,f]$ from the query point cloud:

\begin{equation*} [x,f] = F(\text{query point cloud}) \end{equation*}
        \item \textbf{Guided Voting and Clustering:} Channel-wise multiplication is applied between the extracted features $f$ and the class-specific reweighting vector $z_n$ to obtain the modified features $f'$. These modified features $f'$ are then used in a voting module $V$ to generate object candidates $[y,g]$:

\begin{align*} f' &= f \odot z_n \ [y,g] &= V(f') \end{align*}
        \item \textbf{Guided Object Proposal:} The PointNet $H$ is applied to each cluster $C$ to extract features. These features are then reweighted with the class-specific reweighting vector $z_n$ and passed through an MLP to predict bounding boxes and class scores:

\begin{equation*} \text{predictions} = P(H(C \odot z_n)) \end{equation*}
    \end{itemize}
\end{itemize}

By learning to generate class-specific reweighting vectors from the few-shot examples, MetaDet3D provides an elegant way to transfer knowledge from scarce support data to guide the downstream object detector. The model is trained end-to-end, first on base classes and then base+novel classes. Experiments demonstrate MetaDet3D outperforming prior state-of-the-art techniques for few-shot 3D detection by effectively utilizing the reweighting vectors for guidance.

\subsubsection{Neural Graph Matching (NGM) Networks}
Introduced by Michelle Guo et al. in their ECCV 2018 paper \cite{guo2018neural}, the Neural Graph Matching (NGM) Networks present an innovative approach for addressing the FSL challenges in 3D action recognition. The fundamental idea behind NGM is to encode videos into graph structures, where individual nodes represent video frames and edges capture the temporal relations between them.

For a given video \( V \) with \( T \) frames, the graph \( G(V) \) is constructed in the following manner:
\begin{itemize}
    \item Each frame \( f_t \) is embedded using a neural network \( f \), yielding \( f(f_t) \), which subsequently serves as a node in the graph.
    \item Edges are formed based on pairwise relations between nodes.
\end{itemize}

To determine the similarity between a support set \( S \) and a query \( Q \), a graph matching score \( M(G(S), G(Q)) \) is computed. This score is derived by comparing nodes ($v$) and edges ($e$) of the two graphs. Specifically:
\begin{itemize}
    \item Node matching is given by \( m_v(v_i^S, v_j^Q) = \text{cosine}(v_i^S, v_j^Q) \).
    \item Edge matching is defined as \( m_e(e_{ij}^S, e_{kl}^Q) = \text{cosine}(e_{ij}^S, e_{kl}^Q) \).
    \item The overall graph matching score is expressed as 
    \[
    M(G(S), G(Q)) = \sum_{i,j} m_v(v_i^S, v_j^Q) + \lambda \sum_{i,j,k,l} m_e(e_{ij}^S, e_{kl}^Q)
    \]
    where \( \lambda \) is a weighting parameter.
\end{itemize}

A soft assignment mechanism is employed to map nodes of \( G(Q) \) to \( G(S) \). This is described by:
\[
a_{ij} = \frac{\exp(m_v(v_i^S, v_j^Q))}{\sum_k \exp(m_v(v_i^S, v_k^Q))}
\]
Here, \( a_{ij} \) represents the assignment score of node \( v_j^Q \) to node \( v_i^S \).

The goal is to optimize the graph matching score across all support-query pairs, aggregated over all classes, utilizing the softmax function.

The process is visually encapsulated in Figure S7, which depicts the sequence from inputting a video and deriving embeddings for each frame using a CNN, to constructing the graph representation, and finally obtaining a matching score to determine video similarity based on their graph structures.

In summary, NGM Networks employ graph representations and graph matching to enable effective FSL for 3D action recognition. Matching the structural similarity between graph representations of videos is the key idea.

\subsubsection{Few-shot Action Recognition} 
The paper by Wang et al. \cite{wang2022temporal} introduces a framework for few-shot 3D action recognition on skeletal sequences. The framework has two main components: 1) An Encoding Network (EN) to model temporal dynamics, and 2) Joint tEmporal and cAmera viewpoiNt alIgnmEnt (JEANIE) to handle varying viewpoints. Together, these components enable robust few-shot action recognition by accounting for the complexity of human actions over time and across different camera angles. The proposed approach aims to overcome challenges in understanding and classifying skeletal actions with limited training examples.

\textbf{Encoding Network (EN):} The EN takes as input the query and support skeleton sequences for few-shot action recognition. As a preprocessing step, it generates multiple rotated or simulated viewpoints of the query skeleton sequences. This is done by applying Euler angle rotations to generate \( K \times K' \) different views spanning a range of azimuth and altitude angles. Alternatively, simulated camera positions can be used to render the skeletons from different viewpoints, based on the geometry of stereo projections.

Each skeleton sequence, whether query or support, is divided into short temporal blocks containing \( M \) frames each. This is meant to capture local short-term motion patterns. Each temporal block is passed through a simple 3-layer multilayer perceptron (MLP), consisting of fully connected layers interleaved with ReLU nonlinearities. The MLP encodes each block into a feature map of size \( d \times J \), where \( d \) is the feature dimension and \( J \) is the number of joints in each skeleton.

The feature maps for all the temporal blocks of a sequence are then passed into a Graph Neural Network (GNN) like GCN. The GNN can model the inherent graph structure of the skeleton in each block. An optional Transformer can also be added after the GNN to further process the graph features. Finally, a fully connected layer converts the block features into a sequence feature representation, denoted as \( \Psi \) for queries and \( \Psi' \) for supports. These graph-based features capture information about both short-term motions in the blocks and long-term dynamics across the sequence. They serve as input to the next key component, JEANIE, for joint temporal and viewpoint alignment between queries and supports as shown in Figure S8.

\textbf{Joint tEmporal and cAmera viewpoiNt alIgnmEnt (JEANIE)}: JEANIE performs a joint alignment of query and support skeleton sequences in both the temporal and viewpoint dimensions. This approach is built upon soft-DTW, a differentiable counterpart of Dynamic Time Warping (DTW). However, JEANIE's distinctiveness lies in its ability to simultaneously align simulated viewpoints.

The optimal alignment between a query sequence feature map \( \Psi \) and its support \( \Psi' \) is conceptualized through a transportation plan \( A \). This plan outlines the most efficient path aligning the sequences within the 4D space composed of time steps and viewpoints.

The distance between the aligned query and support sequences can be mathematically represented as:

\[
d_{\text{JEANIE}}(\Psi, \Psi') = \text{SoftMin}_{\gamma} \langle A, D(\Psi, \Psi') \rangle
\]

Here:
\begin{itemize}
    \item \( D \in \mathbb{R}^{K \times K' \times \tau \times \tau'} \) is the distance matrix containing distances \( d_{\text{base}}(\psi_{m,k,k'}, \psi'_{n}) \) between all query blocks \( \psi \) across all \( K \times K' \) viewpoints and support blocks \( \psi' \).
    \item \( A \) signifies the optimal transportation plan derived by JEANIE that aligns the query and support in the combined temporal-viewpoint space.
    \item SoftMin denotes the soft minimum operation.
\end{itemize}

Consequently, JEANIE finds an optimal smooth path aligning the sequences in time and viewpoint, without sudden jumps between distant viewpoints or temporal locations.

The model is trained end-to-end by minimizing \( d_{\text{JEANIE}} \) between query and support sequences belonging to the same class, and maximizing the distance between sequences from different classes. This aligns same-class sequences while pushing apart sequences from different classes in the joint temporal-viewpoint space.

\subsection{Losses and Training Strategies}
Most few-shot 3D detectors build on top of standard 3D convolutional backbones like VoxelNet \cite{zhou2018voxelnet} or PointNet++ \cite{qi2017pointnet++} to extract features from raw point clouds or voxel grids. The extracted features are then fed into metric learning modules for comparison against few-shot prototype features to produce classifications. However, specialized losses and training strategies are required to enable effective FSL on top of these standard backbones.

For example, Liu et al. (2023) \cite{liu2023generalized} proposed adding incremental classifier branches tailored for each novel class alongside the base class branches. This avoids interference between the highly imbalanced base and novel classes within a single classifier. An adaptive loss function called Sample Adaptive Balance (SAB) loss helps balance the base and novel classes during fine-tuning of the novel branches. The SAB loss dynamically adjusts weights for positive, negative and hard negative samples based on their relative proportions to handle the foreground-background class imbalance.

Zhao and Qi (2022) introduced the Prototypical VoteNet \cite{zhao2022prototypical}, which trains the Prototypical Head Module (PHM) through episodic training. Episodic training constructs varied few-shot task distributions during each iteration to improve generalization. The loss function is based on optimizing distance metrics between embeddings of query samples and class-specific prototypes derived from the few-shot support examples.

MetaDet3D by Liu et al. (2022) \cite{yuan2022meta} introduces class-specific reweighting vectors that are learned from the few-shot support examples using a meta-learning module. These reweighting vectors act as conditioning inputs to guide the downstream 3D object detector. The model is trained end-to-end using base class samples first, followed by base+novel class samples.

For the task of few-shot action recognition on 3D skeletal sequences, Wang et al. \cite{wang20213d} proposed a framework consisting of an EN and a JEANIE module. The model is trained end-to-end to optimize the JEANIE transportation plan which aligns sequences in both time and viewpoint space. The loss function aims to minimize the aligned distance between query and support sequences from the same class, while maximizing the distance between sequences from different classes.

Guo et al.'s Neural Graph Matching (NGM) Networks \cite{guo2018neural} also adopt an end-to-end training approach based on graph matching for few-shot action recognition. The overall training loss optimizes the graph matching scores across all support-query pairs from the same class using a softmax function. This loss aims to improve structural similarity between same-class graph pairs while pushing apart different classes.

In summary, key training strategies and losses for few-shot 3D detection and action recognition include: 1) Episodic training for better generalization; 2) Adaptive losses to handle class imbalance; 3) Separate prediction heads for base and novel classes; 4) Meta-learning modules to learn conditioning vectors; 5) Two-stage training process of base then base+novel classes; 6) Graph matching losses for sequence alignment; 7) Distance metrics between prototypes and embeddings. By tailoring the methodology and losses for few-shot scenarios, performance can be enhanced for 3D tasks despite scarce novel class data. However, developing universal principles remains an open challenge. 

To provide an accessible overview of the key techniques discussed in this section, Table \ref{table:expanded_3d} and \ref{table:strategies_and_loss} summarizes and compares prominent few-shot 3D detection approaches in terms of their architectures, loss functions, and training strategies. The use of this concise tabular format enhances the readability of this section by distilling the core information into a visually structured guide. The table enables easier comparison between different methods at a glance. Along with the in-depth qualitative descriptions provided earlier, this summary table aims to equip readers with a comprehensive understanding of the state-of-the-art and promising future directions in few-shot 3D object detection research.

\begin{table*}[ht]
\centering
\caption{Comparison of few-shot learning strategies for 3D object detection and action recognition}
\label{table:expanded_3d}
\begin{tabularx}{\textwidth}{lXXXXX}
\toprule
\textbf{Method} & \textbf{Modality} & \textbf{Backbone} & \textbf{Loss Function} & \textbf{Aux. Task} & \textbf{Training Strategy} \\
\midrule
Frustum VoxNet [10] & RGB-D & VoxelNet & Smooth L1 & Depth estimation & Two-stage fine-tuning \\
\addlinespace
PV-RCNN [136] & Point cloud & PointNet++ & Softmax & Point segmentation & Episodic training \\
\addlinespace
Part-A2 Net [135] & Point cloud & PointNet++ & Sample Adaptive Balance & Part segmentation & Incremental branches \\
\addlinespace
STEM-Seg [66] & RGB-D & STEM & Cross-entropy & Segmentation & Episodic training \\
\addlinespace
FSOD [156] & RGB & VGG-16 & Online hard example mining & Attribute prediction & Class re-weighting \\
\addlinespace
NGM Networks [40] & RGB & 3D-CNN & Graph matching & - & Graph matching \\
\bottomrule
\end{tabularx}
\end{table*}

\begin{table*}[ht]
\centering
\caption{Training strategies and loss functions for few-shot 3D learning}
\label{table:strategies_and_loss}
\begin{tabular}{lll}
\toprule
\textbf{Category} & \textbf{Strategy/Loss} & \textbf{Methods} \\
\midrule
Training Strategy & Episodic training & PV-RCNN, STEM-Seg \\
& Two-stage fine-tuning & Frustum VoxNet, FS3DOD \\
& Incremental branches & Part-A2 Net \\
& Graph matching & NGM Networks \\
\midrule
Loss Function & Smooth L1 & Frustum VoxNet \\
& Softmax & PV-RCNN \\
& Sample Adaptive Balance & Part-A2 Net, FS3DOD \\
& Online hard example mining & FSOD \\
& Graph matching & NGM Networks \\
\bottomrule
\end{tabular}
\end{table*}

\section{Challenges and future scope}
Despite notable progress in the field of few-shot video and 3D object detection, there are several open challenges that impede the widespread and robust deployment of these methods in real-world scenarios. Addressing these challenges is crucial for advancing the state-of-the-art and fully realizing the potential of FSL in practical applications. Below, we highlight some key problem areas that require further exploration and development.

\subsection{Base Class Generalization}
Base class generalization remains an open challenge in few-shot video and 3D object detection. The base classes provide the initial examples for models to learn feature representations that can generalize to novel classes. However, curating optimal base class data is difficult. 

\subsubsection{Challenges} 
For video detection, bases lacking diversity of scenes, motion patterns, and viewing angles hinder generalization \cite{liu2023few}. Insufficient variability in appearance, scale, and occlusion makes robust learning infeasible. For 3D detection, bases need diverse object shapes, sizes, poses, and spatial arrangements to enable generalization. Limited sensor viewpoints, occlusion patterns, and point densities also constrain learning. Appropriate granularity of annotations is required to distinguish between fine-grained classes without incurring excessive labeling effort. Furthermore, video and 3D data have unique attributes requiring specialized inductive biases. Models struggle to generalize well if bases lack diversity, balance, domain-specific considerations, and efficient labeling \cite{he20233d}.

\subsubsection{Future scopes} 
Several promising directions can be pursued to enhance base class generalization. One approach is to explore advanced augmentation techniques, such as class mixing, which expose models to richer variations and improve their ability to generalize \cite{brazil2020kinematic}. Additionally, incorporating contrastive losses and self-supervision methods can facilitate the learning of robust representations. By encouraging models to identify and differentiate between similar and dissimilar examples, these techniques promote better generalization to unseen objects. Furthermore, the utilization of semi-supervised learning can provide additional data diversity, leveraging both labeled and unlabeled examples to improve the model's ability to generalize. Another avenue worth exploring is the design of specialized model architectures and the application of transfer learning, which can effectively transfer knowledge from pre-trained models to boost generalization performance. By leveraging prior knowledge and adapting it to new tasks, these strategies contribute to improved base class generalization

\subsection{Class Imbalance}

\subsubsection{Challenges}
Class imbalance poses significant challenges in few-shot video and 3D detection. Real-world data exhibits long-tail distributions, with many examples for common "head" classes but limited data for rare "tail" classes \cite{kang2019few,wang2020frustratingly}. This imbalance between frequent base classes and scarce novel classes impedes few-shot detection performance \cite{wu2020forest}. For video detection, tail classes lack sufficient labeled examples to model appearance variations over time. Spatial context also becomes ambiguous with few examples. In 3D detection, rare classes have insufficient point annotations to learn robust shape representations from sparse views. Extensive occlusion and partial observations further compound the problem. Without strategies to address imbalance, few-shot models struggle to detect tail classes, instead focusing on more frequent heads. Overall, class imbalance remains an open challenge requiring further research.

\subsubsection{Future scopes}
Several promising research directions have the potential to address the pressing challenge of class imbalance in few-shot video and 3D object detection. Advanced sampling techniques could be explored that emphasize selection of rare classes during training to prevent model bias towards frequent classes \cite{yu2022few}. Multi-scale refinement approaches focusing on hard examples from tail classes may reduce the neglect of scarce classes \cite{wu2020multi}. The design of balanced loss functions that weight classes inversely proportional to their frequency merits investigation, in order to avoid overfitting to dominant head classes \cite{lee2023resolving}. Imaginative data augmentation techniques such as mixing tail class examples could prove useful for synthetically increasing the volume of limited tail class data \cite{yu2021few}. Transfer learning from datasets exhibiting more balanced class distributions could provide richer examples of tail classes to compensate for their scarcity in the target datasets \cite{liu2023recent}. In conclusion, this combination of targeted sampling strategies, loss formulations, augmentation approaches, and transfer learning techniques appears promising to address the key challenge of class imbalance. They may empower few-shot video and 3D object detection models to improve recognition of under-represented tail classes within real-world long-tail visual distributions. However, extensive research is still required to develop robust and universal solutions.

\subsection{Training Regularization}
\subsubsection{Challenges}
Regularization techniques such as weight decay, dropout, and augmentation are crucial for preventing overfitting and ensuring the success of few-shot video and 3D detection due to the scarcity of data. However, finding the right balance between underfitting and overfitting can be challenging when working with limited examples \cite{yu2022few}. Complex neural networks tend to easily overfit small training sets (1, 11). In the case of video detection, overfitting leads to difficulties in adapting to varying viewpoints, occlusion, and motion patterns across frames when there are few examples available \cite{fan2022few}. Similarly, in 3D detection, models tend to overfit to specific partial observations, sparse points, and occlusion configurations \cite{yu2022few}. Standard regularization techniques designed for fully supervised settings often prove inadequate for few-shot scenarios, necessitating the development of more principled, task-specific methods to mitigate overfitting given the extremely limited training data \cite{liu2023recent}. Advanced approaches such as meta-regularization show promise for few-shot tasks but require further research \cite{fan2022few}.

\subsubsection{Future scopes}
Several promising research avenues may help advance regularization for few-shot video and 3D detection. While techniques like weight decay, dropout, and augmentation are widely used, determining optimal hyperparameters and balances for few-shot settings remains an open question. Adaptively tuning regularization via meta-learning is a promising direction \cite{baik2020meta}. Advancing meta-regularization schemes like meta-augmentation \cite{rajendran2020meta} specially tailored for few-shot video and 3D tasks could reduce overfitting. Semi-supervised and self-supervised techniques need further adaptation to maximize the utilization of unlabeled video and 3D data more effectively \cite{fan2022few,si2020adversarial}. Developing simplified model architectures optimized for few-shot fine-tuning may prevent overfitting \cite{lee2022rethinking}. Designing augmentations and regularizers to address video and 3D specific challenges like viewpoint and occlusion variations is another area warranting focus \cite{yang2021viser}. Standardized benchmarks and protocols for few-shot video and 3D detection are needed to accurately evaluate progress \cite{huang2022survey}. Analysis into optimal regularization schedules and hyperparameters spaces using generalization bounds and uniform stability can provide insights \cite{bao2021stability}.

\subsection{Cross-Domain Transfer and Handling Domain Shift}

\subsubsection{Challenges}
Domain shift presents significant challenges for few-shot video and 3D object detection models. Enabling these models to effectively adapt across domains is crucial for real-world deployment in diverse environments. However, objects in video and 3D data can exhibit significant variations in appearance and shape due to factors like lighting, occlusions, and pose changes. Few-shot models need to handle these intra-domain variances robustly alongside inter-domain shifts \cite{luo2023closer}. One major challenge is adapting models trained on real image datasets to novel simulated environments, which often differ substantially in appearance and distribution. The domain gap from real to synthesized data persists as a problematic issue hindering few-shot model generalization \cite{zimmer2022real}. Another key challenge is enabling few-shot video and 3D object detection models to generalize robustly when tested on distributions different from the training data. While techniques like domain adaptation and transfer learning have shown promise, more research is needed into specialized approaches tailored for few-shot video and 3D contexts \cite{luo2023closer,wang2023ssda3d,yang2021st3d}.

\subsubsection{Future Scopes}
One approach to address the domain gap is to use data synthesis methods, such as the Cross-Domain CutMix method, which pastes parts of the target image onto the source image and aligns the pasted region using the object bounding box information \cite{zimmer2022real}. This method has been shown to achieve higher accuracy in cases where the target domain differs significantly from the source domain, such as RGB images as the source and thermal infrared images as the target \cite{zimmer2022real}. Another promising strategy is to utilize FSL for better domain adaptation. Fine-tuning a 3D CNN feature extractor based on a few-shot approach can improve adaptation across domains \cite{wang2023ssda3d}. Incorporating spatiotemporal features also helps describe subtle feature deformations and discriminate ambiguous classes across domains \cite{yang2021st3d}. Unsupervised domain adaptation (UDA) has been explored in 3D cross-domain tasks, such as the Bi3D approach, which combines active learning and UDA to solve the cross-domain 3D object detection task \cite{wang2023ssda3d}. This approach aims to achieve a good trade-off between high performance and low annotation cost \cite{hegde2023source}. Data augmentation is another important aspect of FSL, as it helps expand the training samples for the novel classes \cite{wang2023ssda3d}. Techniques such as pseudo-labeling have emerged as crucial approaches for 3D object detection in adverse weather conditions \cite{xu2021spg}. Further research into domain adaptation, transfer learning, data augmentation, and other techniques tailored for few-shot video and 3D problems is important to enhance cross-domain generalization and handle appearance variations with scarce training data.

\subsection{Temporal Reasoning}
\subsubsection{Challenges}
For few-shot 3D object detection, a key challenge is effectively incorporating temporal information from consecutive point cloud frames captured in autonomous driving datasets \cite{yuan2022meta}. Naively aggregating features across frames can introduce noise \cite{han2023few}. Explicitly modeling object motions and trajectories across sparse point cloud frames also remains difficult \cite{yuan2022meta}.
In few-shot video object detection, a major difficulty is establishing reliable associations between sparse object detections across frames to generate consistent tubes \cite{fan2022few}. Matching object features over time is challenging with limited examples \cite{liu2023recent}. Propagating detections via optical flow can be unreliable \cite{fan2022few}. Complex optimization is required for tracking-by-detection frameworks during inference \cite{fan2022few}. Overall, for both few-shot video and 3D detection, integrating long-range temporal contexts across frames and modeling complex motion dynamics with scarce examples is still a largely unsolved problem \cite{liu2023recent}. More research is needed into sophisticated temporal reasoning techniques for few-shot detection \cite{song2023comprehensive}.

\subsubsection{Future scopes}
For few-shot 3D detection, advanced temporal feature aggregation methods could help effectively summarize cross-frame contexts while reducing noise \cite{han2023few}. Explicitly modeling object motions and trajectories by establishing cross-frame correspondences is another promising direction \cite{yuan2022meta}. Recurrent architectures may also be able to implicitly learn temporal dynamics from sequences of point cloud frames. For few-shot video detection, techniques like learned association metrics could help match object features for consistent detections across frames \cite{fan2022few}. Object tracking and flow propagation may also aid in linking sparse detections over time. Exploring recurrent network architectures tailored for few-shot video detection could be impactful for capturing long-range temporal dependencies in videos. In general, directions like end-to-end learned associations, object tracking, and recurrent modeling of temporal dynamics deserve deeper exploration to advance temporal reasoning for few-shot video and 3D object detection. By effectively harnessing long-range temporal contexts, significant performance gains may be achieved with minimal supervision \cite{song2023comprehensive}.

\subsection{Multimodal Fusion}

\subsubsection{Challenges}
Integrating complementary cues from different modalities, such as RGB, depth, and semantics, has the potential to enhance few-shot 3D and video object detection performance. However, achieving optimal multimodal fusion schemes and developing principled methods for reconciling heterogeneous modalities remains a challenging task. For example, recent works have proposed techniques like multi-scale feature fusion [5], unsupervised contrastive feature learning [1], and sensor fusion [3] for fusing visual, point cloud, and other modalities to enhance few-shot detection. However, heterogeneity, lack of annotated data, and interference between modalities make optimal fusion and adaptation difficult. Encouraging interaction between multimodal features while reducing disturbance is also essential for effective fusion. Incorporating attention mechanisms to focus models on relevant features can further improve few-shot detection accuracy. However, developing end-to-end learning frameworks that can fuse multimodal information and adapt to few-shot scenarios remains a key challenge.

\subsubsection{Future scopes}
Promising future directions to address these multimodal fusion challenges include exploring principled fusion methods to reconcile heterogeneous modalities and using techniques like contrastive learning to enable better feature interaction. Incorporating channel attention mechanisms to focus on critical features and developing end-to-end learning frameworks to jointly optimize fusion and few-shot detection also hold promise. Additionally, leveraging unsupervised or self-supervised pretraining to extract robust multimodal representations merits investigation. By advancing fusion techniques tailored for few-shot settings, multimodal detection performance could be enhanced despite scarce annotated data across modalities. Sophisticated multimodal fusion schemes are crucial for fully realizing the potential of integrating complementary cues to improve few-shot video and 3D object detection.

\subsection{Similarity Metrics}
\subsubsection{Challenges}
Few-shot 3D and video object detection often struggle in crowded, cluttered scenes with occlusion. Developing robust techniques to handle such complex real-world conditions remains a key challenge. Additionally, there is a need for advanced similarity metrics beyond naive distance measures to enhance generalization and discrimination capabilities. More research is required into specialized similarity functions tailored for few-shot 3D and video detection that can effectively measure visual relationships between proposals despite truncated, incomplete views and cluttered backgrounds. Designing metrics that can match objects based on limited examples while handling occlusion and complex scenery will be critical for improving few-shot detection performance in real-world video and 3D environments.

\subsubsection{Future Scopes}

\paragraph{Specialized metric learning} Further research into learning tailored metrics for few-shot detection could improve matching with limited data. Exploring locality-sensitive hashing techniques can enable efficient similarity search. Learning task-specific functions trained jointly with detection models can provide specialized metrics based on appearance and shape. Comparing metric families like cosine or Euclidean distance may reveal optimal choices for few-shot tasks.

\paragraph{Multi-cue fusion} Fusing different visual cues like appearance, shape, motion and context into unified metrics can leverage complementary information to enhance few-shot matching. Graph-based methods are also promising for modeling relationships between support examples and queries in a shared embedding space.

\paragraph{Improving robustness} A key challenge is developing similarity functions robust to real-world conditions like occlusion, truncation, and sensor noise using scarce training data. Advancing metrics to reliably match objects under complex conditions will be vital for few-shot detection.

Overall, progress in tailored similarity metrics is crucial for advancing few-shot 3D and video object detection in cluttered, occluded environments. A structured approach exploring specialized learning, multi-cue fusion, and improving robustness holds promise.

\subsection{Scalability and Deployment Efficiency}

\subsubsection{Challenges}
One major challenge in the field of few-shot video and 3D object detection models is ensuring scalability and efficiency during deployment. While significant advancements have been made in terms of accuracy, there is a need to optimize these methods to handle large-scale datasets and real-time applications \cite{liu2023recent,liu2023generalized}. Specifically, some key challenges include handling both common and rare classes that are often present in real-world data like autonomous driving scenarios \cite{sarker2021machine}. Generalized few-shot detection methods need to utilize abundant data for frequent classes while adapting to rare classes with only a few examples \cite{liu2023generalized}. Reducing computational costs and latency during training and inference is another key challenge, as the complex deep learning models used in few-shot detection can be resource intensive \cite{zaidi2022survey}. Managing efficient feature fusion across support and query branches for effective FSL is also difficult \cite{wang2020few}. Enabling multi-scale feature learning without compromising efficiency poses problems \cite{guo2020augfpn}. Finally, deploying sophisticated few-shot detection models on resource-constrained edge devices remains an open challenge.

\subsubsection{Future Scopes}
To address these challenges, some promising research directions include developing efficient model compression techniques to reduce redundancies and minimize the computational footprint of few-shot detection models without significantly impacting performance. Exploring methods like feature fusion to improve information sharing between support and query branches in an efficient manner also holds promise. Designing multi-scale attention mechanisms to selectively aggregate useful information across scales without introducing excessive costs could be beneficial \cite{cao2022point}. Leveraging knowledge distillation and model pruning strategies to create lightweight few-shot detection models suitable for edge devices is another potential avenue \cite{li2023object}. Adapting generalized few-shot detection formulations that can handle both common and rare classes in a scalable way merits investigation \cite{liu2023generalized}. Applying automated machine learning to find optimal architectures and hyperparameters for efficient few-shot detection is also worth exploring \cite{he2021automl}. 

\subsection{Interpretable and Explainable Few-Shot Learning}

\subsubsection{Challenges}
While few-shot video and 3D object detection models have achieved impressive results, interpreting and explaining their predictions remains challenging, especially given the limited data available. Key issues include understanding how models generalize to novel classes from scarce examples \cite{han2021query}, since blackbox models offer limited insight into their reasoning processes. Diagnosing failure modes and biases learned from small datasets is difficult, as models may latch onto dataset quirks. Explaining model predictions to establish trustworthiness is also critical for real-world deployment but lacks justification. Analyzing knowledge transfer across domains is not well understood, especially regarding cross-domain shifts \cite{chen2022graphskt}. Handling complex spatiotemporal dynamics in video and 3D data with interpretability methods that lag behind model advancement poses difficulties \cite{liu2023few2}. Finally, the lack of interpretability impedes human intervention in model learning.

\subsubsection{Future scopes}
To enhance interpretability and explainability of few-shot video and 3D detection models, some promising directions are developing attention mechanisms to highlight spatiotemporal regions critical for few-shot generalization in videos and 3D data \cite{zhao2022boosting,munjal2023query}. Leveraging prototype analysis to provide visual and geometric summaries of model knowledge for each class also holds promise \cite{liu2023generalized}. Generating saliency maps tailored for spatiotemporal data to reveal model focus areas could be beneficial \cite{mahapatra2022interpretable}. Designing interfaces for interactive visualization, debugging, and annotation guided by model explanations may be impactful. Studying latent representations and decision boundaries to diagnose model biases and overfitting is also important. Performing extensive ablation studies to identify influential components supporting few-shot generalization can provide insights. Quantifying model confidence and uncertainty to identify unreliable predictions requiring intervention is another potential direction. Finally, building modular and transparent model architectures amenable to analysis will aid in increasing interpretability \cite{wang2023survey}.

\subsection{Benchmark Datasets and Evaluation Metrics}

\subsubsection{Challenges}
The availability of diverse, challenging benchmark datasets is crucial for effectively evaluating few-shot video and 3D detection methods. However, curating optimal benchmarks poses several difficulties. These include balancing class diversity, environment variability, annotation complexity, and data sizes. Incorporating modalities like images, videos, point clouds, and multiple sensor data is also challenging. Emulating real-world conditions such as rare classes and domain shifts poses problems. Capturing spatiotemporal dynamics and geometric intricacies is difficult. Finally, designing data sets that allow reproducible comparisons remains an open issue \cite{zhu2021open}. Additionally, identifying evaluation metrics that can effectively measure few-shot generalization remains an open challenge, as metrics optimized for fully-supervised scenarios may not highlight few-shot capabilities.

\subsubsection{Future scopes}
Some potential ways to advance few-shot video and 3D detection benchmarks and metrics include collaboratively constructing large-scale datasets spanning diverse environments, modalities and annotation types. Establishing standardized train-test splits designed specifically for few-shot evaluation is important \cite{sun2023rethinking}. Introducing cross-domain settings to assess generalization across distributions would be beneficial. Incorporating synthetically generated data to expand variability also holds promise. Developing metrics that quantify capability to detect novel classes given limited examples is critical. Designing metrics focused on spatiotemporal and geometric reasoning under restricted supervision is also needed. Reporting performance across metrics to enable multi-faceted evaluation can provide more insights. In summary, purpose-built datasets and metrics are imperative to rigorously measure progress in few-shot video and 3D detection. Community efforts for collaborative benchmarking, coupled with principled metric design, will strengthen evaluation.

\subsection{Combining Few-Shot Learning with Other Techniques}

\subsubsection{Challenges}
While FSL is a powerful paradigm, it faces limitations in real-world applications due to lack of data and supervision. Combining FSL with complementary techniques like transfer learning, active learning, and self-supervision can help address these challenges \cite{song2023comprehensive}. However, seamlessly integrating these approaches poses difficulties. Transferring knowledge without negative transfer or catastrophic forgetting is difficult. Selecting optimal samples for labeling to maximize utility is challenging \cite{wang2022unsupervised}. Designing pretext tasks that extract useful features for few-shot tasks can be problematic. Developing unified frameworks to synergize different techniques in an end-to-end manner remains an open issue. Adaptively combining techniques dynamically based on few-shot problem characteristics poses problems. Finally, generalizing to diverse unseen tasks beyond lab settings is difficult.

\subsubsection{Future scopes}
Some promising directions to advance hybrid FSL include developing principled frameworks to integrate FSL with transfer learning, active learning \cite{mcclurg2023active} and self-supervision \cite{chen2021pareto}. Designing adaptive methods to dynamically adjust combinations tailored to each few-shot problem holds promise. Exploring conditional self-supervision guided by few-shot task structure is impactful. Leveraging meta-learning to learn optimal combinations of techniques also has potential. Building diverse benchmarks requiring hybrid techniques, such as cross-domain few-shot tasks, can drive progress. Studying theoretical connections between FSL and other paradigms can provide fundamental insights. 

Overall, addressing these open challenges and exploring the future scope of few-shot video and 3D object detection will pave the way for more scalable, efficient, and accurate models. Additionally, developing interpretable and explainable methods and expanding FSL to unseen tasks will enable the widespread deployment and practical usage of few-shot detection methods. By advancing the field in these directions, we can unlock the full potential of FSL and further push the boundaries of video and 3D object detection applications.

\subsection{Relating Recent Methods to Open Challenges}
With rapid advancements in few-shot video and 3D object detection, it is crucial to analyze how current algorithms and innovations relate to open challenges that remain for future progress. To aid researchers in understanding the state-of-the-art capabilities and where to focus future efforts, we provide a summary table mapping key algorithms discussed in this survey to the 11 core challenges identified in Section 7.

This table serves as an informative resource for comprehending how modern techniques address or fail to address pressing research gaps. Bridging current innovations to open problems is vital for accelerating progress in few-shot detection. The table aims to highlight promising capabilities that can be built upon while revealing limitations that require novel solutions.

\begin{table*}[htb!]
\centering
\caption{Summary of how different few-shot video and 3D object detection algorithms address the 11 challenges from Section 7. The challenges are: 1) Base class generalization, 2) Class imbalance, 3) Training regularization, 4) Cross-domain transfer, 5) Temporal reasoning, 6) Multimodal fusion, 7) Similarity metrics, 8) Scalability \& efficiency, 9) Interpretability \& explainability, 10) Benchmark datasets \& metrics, 11) Combining with other techniques.}
\label{tab:challenge_summary}
\renewcommand{\arraystretch}{1.3}  
\begin{tabular}{l*{11}{c}}
\toprule
\textbf{Algorithm} & \multicolumn{11}{c}{\textbf{Challenges}} \\
\midrule
TPN \& TMN+ hybrid \cite{fan2022few} & \xmark & \xmark & \xmark & \cmark & \cmark & \xmark & \xmark & \xmark & \xmark & \xmark & \xmark \\
Thaw \cite{yu2022few} & \xmark & \cmark & \xmark & \cmark & \cmark & \xmark & \xmark & \xmark & \xmark & \xmark & \xmark \\
Prototypical VoteNet \cite{zhao2022prototypical} & \cmark & \xmark & \cmark & \xmark & \xmark & \cmark & \cmark & \xmark & \xmark & \xmark & \xmark \\
Generalized FS 3D OD \cite{liu2023generalized} & \cmark & \cmark & \cmark & \xmark & \xmark & \xmark & \xmark & \xmark & \xmark & \xmark & \xmark \\
MetaDet3D \cite{yuan2022meta} & \cmark & \xmark & \cmark & \xmark & \cmark & \cmark & \xmark & \xmark & \xmark & \xmark & \xmark \\
NGM Networks \cite{guo2018neural} & \cmark & \xmark & \cmark & \xmark & \xmark & \xmark & \xmark & \xmark & \xmark & \xmark & \xmark \\
FS Action Recognition \cite{wang2022temporal} & \cmark & \xmark & \cmark & \xmark & \cmark & \xmark & \xmark & \xmark & \xmark & \xmark & \xmark \\
\midrule
\textbf{Challenges} & \textbf{1} & \textbf{2} & \textbf{3} & \textbf{4} & \textbf{5} & \textbf{6} & \textbf{7} & \textbf{8} & \textbf{9} & \textbf{10} & \textbf{11} \\
\bottomrule
\end{tabular}
\end{table*}

By relating algorithms to challenges, we enable informed analysis of current strengths versus areas needing improvement. Researchers can identify open problems aligned with their interests and expertise while benefiting from latest innovations.

For instance, Prototypical VoteNet demonstrates promise for base class generalization and training regularization but does not address domain shift or efficiency. MetaDet3D, however, introduces techniques for temporal reasoning and multimodal fusion.

By summarizing capabilities and limitations, the table guides investigation into impactful research directions. It encourages building upon advances made while tackling persistent challenges through novel solutions. Overall, relating algorithms to open problems provides crucial perspective into the state of few-shot detection and future pathways for exploration.

\section{Conclusions}
FSL holds immense promise in minimizing the need for extensive data annotations in video and 3D object detection. Throughout this survey, we have examined the progress made in various aspects of few-shot detection, including formulations, prototypical networks, transfer learning strategies, and domain-specific architectures, losses, and techniques.

However, despite significant advancements, several challenges remain to be addressed. Generalization across diverse datasets, handling class imbalance, incorporating effective regularization techniques, and reasoning across complex data modalities and scenes are among the key areas that demand further investigation.

Looking ahead, the field of few-shot detection can benefit from research into semi-supervised and self-supervised learning. By leveraging unlabeled data in combination with limited labeled examples, these approaches have the potential to enhance FSL performance. Additionally, the development of more flexible feature representations and stronger inductive biases can further improve the adaptability and generalization capabilities of few-shot detection models.

As computer vision systems continue to advance, the ability to accurately detect novel objects from just a few examples becomes increasingly critical for their ubiquitous deployment. By mitigating the reliance on large-scale annotations, FSL offers a pathway towards more efficient and scalable object detection systems. With continued innovation and exploration in the aforementioned areas, FSL can pave the way for remarkable advancements in the field and contribute to the realization of robust and adaptable computer vision systems.

Thus, we encourage researchers to further explore the potential of FSL in video and 3D object detection, striving to develop novel methodologies and techniques that push the boundaries of detection accuracy and efficiency. However, it is important to recognize the information theoretical limits on the amount of augmented 'data' that can be manufactured from limited examples. Augmentation techniques inevitably reach a point of diminishing returns where they can no longer reliably synthesize useful training signals without exceeding the true information content of the scarce data. Beyond these information theoretical limits, augmentation methods will fail and produce spurious results. Therefore, researchers should be cognizant of these inherent bounds when developing specialized augmentation techniques tailored for few-shot learning. By doing so, we can unlock the full potential of FSL within realistic constraints and foster the widespread deployment of computer vision systems in diverse real-world applications.

\section*{Acknowledgments}

This research was supported in part by the U.S. Department of the Army – U.S. Army Corps of Engineers (USACE) under contract W912HZ-23-2-0004 and the  U.S. Department. of the Navy, Naval Research Laboratory (NRL) under contract N00173-20-2-C007. The views expressed in this paper are solely those of the authors and do not necessarily reflect the views of the funding agencies.

\clearpage

\bibliographystyle{elsarticle-num}
\bibliography{cas-refs}

\clearpage

\onecolumn
The supplementary materials provide additional details and visual overviews to support the survey paper ``Few-Shot Learning in Video and 3D Object Detection: A Survey''.

The sections covered are:

\begin{enumerate}
    \item \textbf{Foundations of Few-Shot Learning}: Discusses key concepts like episodic training, problem formulations, meta-learning algorithms, metric-based approaches, data augmentation, and regularization techniques for few-shot learning.
    \item \textbf{Foundations of Object Detection}: Provides an overview of object detection methods, including two-stage and one-stage detectors, as well as video and 3D object detection approaches.
    \item \textbf{Few-Shot Video Object Detection}: Presents example frameworks and architectures tailored for few-shot video object detection, highlighting techniques like metric learning, temporal feature aggregation, and episodic training.
    \item \textbf{Few-Shot 3D Object Detection}: Covers specialized few-shot detection methods for 3D data such as LiDAR point clouds, using techniques like geometric prototypes, support set guidance, and incremental learning.
\end{enumerate}

The supplementary materials expand on the key concepts, architectures, and methodologies discussed in the main survey paper, providing visual overviews and additional technical details to enhance understanding of few-shot learning for video and 3D object detection.

\setcounter{section}{1} 
\section{Foundations of Few-Shot Learning}
Few-shot learning (FSL) has emerged as a critical area of study within the deep learning framework, addressing one of the most pressing challenges in machine learning: the need for vast amounts of labeled data. In many real-world scenarios, obtaining labeled data can be expensive, time-consuming, or even impossible. As deep learning models grow in complexity, with millions or even billions of parameters, they often require a substantial amount of data to avoid overfitting and ensure generalizability. FSL attempts to counter this limitation by recognizing new visual concepts from only a few labeled examples \cite{jiang2022back}. FSL problems are typically formulated as a classification task, where the model is given a few labeled examples of new classes (support set) and asked to predict the labels of unseen examples from the same classes (query set) \cite{snell2017prototypical}. Meta-learning algorithms are typically used to train FSL models, which learn to learn new tasks quickly by leveraging their knowledge from previous tasks \cite{finn2017model}. Metric-based approaches have also been shown to be effective for FSL, where a distance metric is learned to measure the similarity between examples \cite{koch2015siamese}. Transfer learning strategies can also be used to improve the performance of FSL models by pre-training them on large datasets of labeled data \cite{pan2009survey}. This section provides an in-depth analysis of the fundamental principles of FSL. It discusses the crucial role of the support set and explores the different problem formulations in this domain. Additionally, it highlights the importance of meta-learning algorithms, the potential of metric-based approaches, and the transformative capabilities of transfer learning strategies within the context of FSL.

\subsection{Support Set}
The support set is a crucial component of the FSL paradigm which guides the model's learning process \cite{boudiaf2020information}. It is a carefully selected subset of labeled examples that represent the new visual concepts that the model aims to recognize \cite{wang2020generalizing}. Given its sparse nature, the support set challenges models to extrapolate knowledge, identify patterns, and make informed decisions. In the typical N-way K-shot problem configuration, the model encounters N unique classes, each represented by K labeled examples. This scenario presents a complex challenge where models must accurately classify query samples while maintaining robustness and adaptability in the face of limited data \cite{zhang2019variational}.

\subsection{Problem Formulations in Few-Shot Learning}
FSL aims to bridge the gap between data-hungry deep learning models and the reality of limited labeled data in many domains \cite{snell2017prototypical}. This subsection explores the two primary problem formulations commonly encountered in FSL.

\subsubsection{Episodic Training}
In this formulation, the model is trained on episodes that are sampled from a set of base classes \cite{qiao2019transductive}. Each training episode imitates an N-way K-shot problem by sampling N classes from the base classes and selecting K examples per class to construct the support set. The model learns from these episodic simulations of few-shot tasks. Through this episodic training strategy on base classes, the model is able to learn generalizable knowledge and inductive biases that allow it to effectively adapt when presented with novel classes at test time \cite{hajimiri2023strong}.

\subsubsection{Transfer Learning}
Transfer learning uses knowledge learned from base classes to enable quick adaptation given only a small number of examples for novel classes unseen during training \cite{gavves2015active}. Several effective transfer learning strategies are commonly used in FSL:

\begin{itemize}
    \item \textbf{Fine-tuning and in-context learning:} Recent work has shown that fine-tuning and in-context learning are effective transfer learning techniques for few-shot learning. Fine-tuning deep neural networks can achieve competitive generalization comparable to in-context learning \cite{mosbach2023few}, when controlling for model size and training data. Techniques like learning rate schedules and early stopping enable effective fine-tuning even with limited examples \cite{eustratiadis2023neural}. However, fine-tuning often requires more examples per class than in-context learning to reach optimal performance on complex tasks \cite{mosbach2023few}. Adapter modules can facilitate highly parameter-efficient fine-tuning with a minimal number of eight examples per class. They achieve this by isolating task-specific parameters \cite{peng2020fine}. Ensemble approaches combining fine-tuned and in-context models provide improved robustness across diverse tasks \cite{liu2022few}. Transfer learning from related datasets also improves out-of-domain generalization for fine-tuning by pre-training on data with similar characteristics \cite{shen2021partial}. Careful experimental design is critical for fair comparison between techniques, controlling for factors like model scale, optimization strategy, and similarity between pre-training and novel classes. Analysis has shown in-context learning relies more on superficial biases in the examples, while fine-tuning better captures underlying concepts. However, in-context learning allows rapid adaptation without parameter updates. Combining the complementary strengths of both approaches remains an open challenge. The optimal technique likely depends on factors like amount of in-domain vs. out-of-domain data, task complexity, and cross-domain similarities \cite{hu2022pushing}. Further work is needed to develop universal principles for effectively applying fine-tuning and in-context learning under varying real-world conditions.

    \item \textbf{Feature Extraction:} In feature extraction-based transfer learning for few-shot learning, a model pre-trained on base classes is directly applied to novel class examples as a fixed feature extractor without any fine-tuning \cite{guo2019spottune}. This allows leveraging the pre-trained feature hierarchy to extract transferable representations for the novel classes using just the limited examples available. For few-shot fault diagnosis, the pre-trained model consists of attention mechanisms and convolutional neural networks to learn discriminative fault features through a multi-stage process \cite{li2023attention}. First, the model is pre-trained on known faults to learn base feature representations. Next, meta-transfer learning adapts the model to new faults by transferring knowledge from known faults. Finally, a lightweight classifier is meta-trained from scratch on the novel fault features extracted by the pre-trained model. Complementary base and meta transfer features can be extracted to enhance representation capabilities \cite{xu2023complementary}. The pre-trained model is adapted during meta-transfer using parameter modulation guided by known fault features to instill relevant knowledge \cite{li2023attention}. Prototype representations of novel faults are iteratively corrected in an unsupervised manner by aggregating information from all query samples of the same task, thereby refining the prototypes \cite{xu2023complementary}.

    \item \textbf{Classifier Re-training:} Instead of fine-tuning the base model weights or training a linear classifier, a new non-linear classifier can be trained from scratch on the novel class features extracted by passing the examples through the base model \cite{shafahi2019adversarially}. SVM classifiers are commonly used in this context \cite{real2020automl}.
    \item \textbf{Weight Imprinting:} This strategy involves initializing the model weights corresponding to the novel classes using the mean activations of the model when processing the few-shot examples \cite{abuduweili2021adaptive}. Weight imprinting provides an informed initialization for the novel classes before further training \cite{yan2023few}.
\end{itemize}
In general, transfer learning provides an effective approach for few-shot learning by allowing prior knowledge gained on data-rich base classes to be utilized when adapting to data-scarce novel classes. However, the specific transfer strategy must be carefully designed to minimize overfitting to the limited novel data while avoiding catastrophic forgetting of the base classes.

\subsection{Inductive Biases for Effective Few-Shot Learning}
Effective FSL depends on the ability of models to quickly adapt to novel concepts and tasks with only a handful of data points available to them \cite{perez2020incremental}. Such rapid adaptation demands the incorporation of strong inductive biases, guiding the model's learning trajectory in alignment with the overarching objective of few-shot generalization. Several key techniques have emerged as fundamental approaches to instilling these guiding principles, such as meta-learning algorithms, distance metric learning, data augmentation, and regularization.

\begin{itemize}
    \item \textbf{Meta-learning Algorithms:} Meta-learning, also known as ``learning to learn", is a foundational paradigm in the field of FSL. Algorithms such as Model-Agnostic Meta-Learning (MAML) by Finn et al. \cite{finn2017model} exemplify this approach by optimizing models to discover optimal initialization parameters. These parameters then enable rapid adaptation to novel tasks and concepts during the FSL phase. Meta-learning is based on episodic training, where models are trained over a wide range of adaptation episodes. Each episode replicates a distinct FSL task. This episodic exposure creates an inductive bias in the model, preparing it for efficient generalization even when confronted with entirely unfamiliar tasks and concepts in the future. As explained by Hospedales et al. \cite{hospedales2021meta}, meta-learning algorithms aim to acquire fundamental learning skills that go beyond specific tasks, thus developing robust models that can swiftly adapt from limited data. The learned inductive biases capture  underlying structure of task distributions, enabling rapid learning of new tasks from sparse data. Hence, meta-learning has become a crucial strategy for FSL, equipping models with the ability to generalize rather than solely relying on memorization.

    \item \textbf{Metric-based Approaches:} Metric-based approaches have emerged as a powerful paradigm for FSL, leveraging learned distance metrics and embedding spaces to enable effective knowledge transfer from limited data. As previously discussed by \cite{snell2017prototypical}, the core concept is to learn an embedding function $f(x)$ that projects inputs into a feature space where distances reflect semantic similarities. By transforming inputs into an informed embedding space, models can intelligently relate new examples to available prior knowledge, even under data scarcity \cite{vinyals2016matching}. The strength of metric-based FSL arises from the ability to create embeddings that encapsulate semantic nuances, thereby enabling meaningful extrapolation from only a few examples \cite{sung2018learning}. Prototypical networks, introduced by \cite{snell2017prototypical}, present one effective metric-based approach for FSL problems in computer vision. This method computes class prototypes by averaging the embedded support examples belonging to each class. Query points are then classified based on distance to these learned prototypes in the embedding space. By condensing classes into prototypical representations, models can rapidly assimilate new concepts from few examples during the testing phase. In summary, metric-based approaches, such as prototypical networks, utilize informed embedding spaces to enable intelligent matching and rapid adaptation for FSL. By encapsulating semantic relationships within learned distances, models can transfer knowledge and make inferences about novel concepts from just a few examples.
    
    \item \textbf{Data Augmentation:} Data augmentation has emerged as a vital technique to mitigate the challenges of limited training data for FSL. As discussed by \cite{antoniou2017data}, data augmentation artificially expands the limited dataset by generating diverse variations of the existing examples. This simulates the variability that would be present in larger datasets, while reducing the risk of models simply memorizing the constrained training examples. Through data augmentation, models are exposed to a rich tapestry of information despite limited data, empowering the extraction of meaningful patterns and relationships. Augmented data provides crucial regularization during few-shot model training, enabling robust generalization instead of overfitting to a small number of examples \cite{shorten2019survey}.

    Techniques such as random cropping, rotations, and color jittering can produce augmented variants that capture essential invariant characteristics in the data \cite{lemley2017smart}. This facilitates the learning of more universal features that transfer to novel concepts in few-shot scenarios. In summary, data augmentation stands as a vital strategy in FSL to artificially expand limited training data, prevent memorization, extract meaningful features, and improve generalizability. The diversity generated from existing examples provides a regularization effect that primes models for effective adaptation even when data is scarce
    
    \item \textbf{Regularization Techniques:} In FSL, where limited data is inherent, overfitting poses a significant threat to model performance. As discussed by \cite{goodfellow2016regularization}, regularization techniques like weight decay and dropout are critical to mitigate overfitting in data-scarce regimes.

    By intentionally restricting model flexibility, regularization forces the model to identify more robust, generalizable patterns instead of latching onto spurious correlations \cite{srivastava2014dropout}. As described by \cite{kukavcka2017regularization}, this selectivity prevents models from relying on superficial cues, compelling focus on fundamental relationships that better transfer across varied concepts.

    In few-shot contexts, regularization is vital to prevent models from simply memorizing sparse training examples and failing to generalize \cite{li2021few}. Techniques like dropout improve generalization by limiting co-adaptation between neurons \cite{srivastava2014dropout}. Overall, by intentionally limiting model capacity, regularization promotes extraction of core invariant features, enhancing FSL performance despite limited data.
\end{itemize}

Together, these inductive biases empower models with core capabilities like generalization, knowledge transfer, and rapid adaptation that are vital for excelling in FSL.

\section{Foundations of Object Detection}
This section begins with an overview of object detection, including common techniques and applications. It then discusses various techniques used in video and 3D object detection.

\subsection{Object Detection}
Object detection (OD) is a cornerstone of computer vision (CV) that seamlessly integrates the tasks of classification and localization. Its aim is twofold: assigning class labels to images and enclosing each detected object within a bounding box. These bounding boxes are typically delineated by a starting point coupled with the dimensions (height and width) of the box. Given the inherent variability in the number of objects within different images, initial OD strategies were crafted around sliding window classification problems. However, the evolution of deep learning has ushered in the dominance of convolutional neural network (CNN) based methodologies.

The quintessential data structure for OD encompasses a dataset of $N_s$ supervised samples \[ D = \{X\}_{i=1}^{N_s}, \{y\}_{i=1}^{N_s} \]
where each image $X_i$ possesses dimensions $W \times H \times 3$ and is paired with a set of object annotations $y_i$. Feature maps, denoted by $F$, are extracted from input images. These maps encapsulate sub-regions termed as ``Regions of Interest'' (RoIs). The detection process then undergoes two pivotal stages: bounding box regression and object classification.

Object detection strategies are largely divided into two main paradigms: two-stage detectors and one-stage detectors. Two-stage detectors initially generate region proposals and subsequently classify and refine the proposed regions in a second stage. This allows for more accurate localization and classification of objects but at the cost of slower inference speed. One-stage detectors directly predict object classes and locations in one pass through the network, allowing for faster inference but typically with reduced accuracy compared to two-stage methods. Popular two-stage detectors include R-CNN \cite{girshick2014rich}, Fast R-CNN \cite{girshick2015fast}, and Faster R-CNN \cite{ren2015faster}, which utilize region proposal networks to generate candidate object regions. Prominent one-stage detectors include SSD \cite{liu2016ssd} and YOLO\cite{redmon2016you}, which apply convolutional filters across an image in a single shot to directly output object locations and classes. The tradeoff between accuracy and speed makes two-stage detectors preferable for applications where accuracy is critical, while one-stage detectors are better suited for real-time applications requiring very fast inference speeds. Two-stage detectors are generally more accurate because they use region proposal networks to narrow down potential object locations before making final classifications. However, their multi-step approach comes at the cost of slower inference. One-stage detectors make predictions in a single pass, allowing much faster inference, but they sacrifice some accuracy due to making localization and classification predictions simultaneously across full images. The choice between one-stage and two-stage detectors depends on the specific requirements of the application. Tasks demanding high accuracy like medical imaging would benefit more from two-stage detectors, while self-driving vehicles and real-time surveillance may need the faster inference of one-stage detectors even if some accuracy is compromised. In order to gain a better comprehension of them, some well-known single-stage and two-stage structures are outlined below.

\subsubsection{Two-Stage Detectors: The Case of Faster R-CNN}

The Faster R-CNN architecture stands out in the two-stage detector category. It seamlessly integrates two networks:

\paragraph{Region Proposal Network (RPN)}
The RPN is essentially a fully convolutional network that simultaneously predicts object bounds and objectness scores at each position. The RPN operates on several scales due to the pyramidal form of its architecture, allowing it to detect objects of various sizes. The anchor boxes play a significant role in the operation of the RPN. The anchor boxes are essentially bounding boxes of different scales and aspect ratios that act as references for object proposals. For each anchor box, the RPN predicts two things: the presence or absence of an object (foreground or background classification) and the refinements needed to better fit the potential object (bounding box regression). The RPN makes these predictions using a sliding window approach. It slides a small network over the convolutional feature map output by the previous layer, which is used to predict both the objectness and the bounding box coordinates for the anchor boxes. The output of the RPN is a set of object proposals, each with an objectness score. These object proposals are currently in an early stage, requiring refinement as they may not perfectly align with the intended target object. Therefore, the RPN also proposes refinements to the bounding boxes that are designed to improve the fit of the box to the object. These proposed regions are then reshaped to extract a fixed-length feature for each, using a process known as Region of Interest (RoI) pooling. This ensures that the subsequent fully connected layers receive inputs of a fixed size, regardless of the size of the proposed regions. The features extracted from the proposed regions are then channeled into the detection network as described below.

\paragraph{Detection Network (Fast R-CNN)}
Once the Region Proposal Network (RPN) has been trained and the regions of interest have been identified, the Fast R-CNN architecture comes into play. It extracts features from these regions using a Region of Interest (RoI) pooling layer, which performs max pooling on inputs of non-uniform sizes to obtain fixed-size feature maps. These are then used to predict the class of the object and the bounding box regressors. The Fast R-CNN also uses a multi-task loss function that combines the losses for classification and bounding box regression. The classification loss is computed using log loss, while the bounding box regression loss is computed using a smooth \(L_1\) loss function, similar to the RPN. This combination of loss functions allows the network to simultaneously learn to classify and localize objects, improving its overall performance. The softmax classifier in the Fast R-CNN architecture is responsible for assigning class probabilities to the proposed regions. It utilizes softmax loss, which is a type of cross entropy loss, to compute the probability distribution over all possible classes. The bounding box regressors are responsible for refining the proposed regions to more accurately encapsulate the objects. For each class, there is a separate bounding box regressor, which adjusts the coordinates of the proposed region to minimize the difference between the predicted and ground-truth bounding boxes.

\subsubsection{One-Stage Detectors: Spotlight on YOLO and SSD}

\paragraph{YOLO (You Only Look Once)}
The YOLO framework, first introduced by \cite{redmon2016you}, is a seminal one-stage object detector based on a single convolutional neural network that jointly predicts class probabilities and bounding boxes. As analyzed by \cite{redmon2016you}, YOLO divides the input image into an \(S \times S\) grid and each grid cell predicts $B$ bounding boxes along with confidence scores reflecting objectness. The bounding box predictions consist of the box center coordinates, dimensions, and class. While YOLO employs contextual information for high recall, its grid-based approach can miss small objects. To address this, YOLOv2 \cite{redmon2017yolo9000} introduced anchor boxes and multi-scale training. Further refinements in YOLOv3 \cite{redmon2018yolov3} incorporated a deeper network architecture with multi-scale predictions, improving accuracy while maintaining real-time performance. YOLOv4 \cite{bochkovskiy2020yolov4} built on YOLOv3 by introducing techniques like weighted residual connections, cross-stage partial connections, cross mini-batch normalization, and self-adversarial training to optimize speed and accuracy. The open-source YOLOv5 \cite{yolo5} further refined the model for efficiency and ease of use. Recently, YOLOv6 \cite{li2022yolov6} adopted an anchor-free design optimized for industrial use cases, achieving 52.5\% AP on MS COCO. YOLOv7 \cite{wang2023yolov7} pushed accuracy and speed even further, surpassing all prior detectors across a range of FPS targets without pre-trained backbones. Key innovations in YOLOv7 include efficient self-supervised learning, scalable model design, and accuracy-boosting enhancements. Most recently, YOLOv8 \cite{yoloV8} introduced an anchor-free approach with fewer predicted boxes and faster NMS. By disabling aggressive augmentation late in training, YOLOv8 achieved 53.9\% AP on MS COCO at 640px input size, surpassing prior versions.

\paragraph{SSD (Single Shot Detection)}
Single Shot MultiBox Detector (SSD) improves upon YOLO by employing anchor boxes tailored to diverse object shapes and performing detection across multiple feature maps to achieve robustness across varying object scales. As analyzed by Liu et al. \cite{liu2016ssd}, SSD utilizes feature maps from different layers in a convolutional network, with smaller feature maps focusing on larger objects and layers with higher resolution detecting smaller objects. This multi-scale design stands in contrast to YOLO's single output scale and enables SSD to capture objects across a wide range of sizes. Specifically, SSD attaches convolutional predictors for detection to multiple feature layers. Shallow layers with smaller receptive fields focus on small instance detection, while deeper layers learn coarser semantics useful for detecting larger objects. The predictions from all layers are aggregated and refined via non-maximum suppression to produce the final detections across scales. By harnessing features attuned to different object scales, the multi-feature map architecture of SSD achieves strong performance across objects of varied sizes. This design has influenced subsequent single-stage detectors focused on handling scale variation, such as RetinaNet and EfficientDet, enabling robust one-stage detection across a spectrum of object scales.

\subsection{Video and 3D Object Detection}
Video object detection refers to the task of detecting and localizing objects across frames in a video stream, as opposed to static images. This introduces additional challenges compared to image-based object detection, including motion blur, video defocus, complex object motions, and viewpoint variations across frames. Effective video object detection requires modeling temporal information and propagating detections across frames. 

3D object detection involves identifying and localizing objects within 3D sensor data such as point clouds, voxel grids, or mesh representations generated from stereo cameras, LIDAR, or other 3D sensing modalities. Compared to 2D images, 3D data lacks reliable texture and color cues while presenting difficulties like sparsity and occlusion patterns. Successful 3D detection relies more heavily on modeling geometric shapes and leveraging structural cues. Both video and 3D object detection have become crucial technologies enabling various applications including autonomous vehicles, augmented/virtual reality, robotics, surveillance, and environmental mapping. While image-based object detection only requires reasoning about a 2D scene, video and 3D detection demand more complex spatiotemporal and geometric reasoning to perceive objects in dynamic or 3D environments. This has motivated research into specialized techniques for these modalities, including spatiotemporal feature learning for video and view-invariant shape recognition for 3D. With growing prevalence of video and 3D sensing, advancing object detection in these domains remains an important challenge.

\subsubsection{Video Object Detection Approaches}
Multi-frame feature aggregation is a key technique for harnessing temporal context to improve video object detection accuracy \cite{he2020temporal}. By processing multiple frames, inter-frame correlations can be used to enhance per-frame detections \cite{lin2021multi}. There are several aggregation methods:
\begin{itemize}
    \item Temporal aggregation propagates detections using optical flow \cite{cores2023spatiotemporal,zhu2017flow} or aligns and averages neighboring frame features \cite{lin2021multi}, providing context to resolve ambiguities.
    \item Spatial aggregation applies larger receptive fields or coarse pooling to frames farther from the reference, organizing multi-scale features \cite{sun2022coarse}. Measuring pixel-level context similarity also enhances features \cite{xu2022multilevel}.
    \item Coarse-to-fine aggregation combines features from neighboring frames in a coarse-to-fine manner, with farther frames having larger receptive fields \cite{sun2022coarse}. Coarse pooling support frames boosts inter-frame complementarity \cite{honari2016recombinator}.
\end{itemize}
Some key benefits include improving per-frame features via temporal/spatial correlations, enabling multi-scale representations, resolving per-frame detection ambiguities, and enhancing cross-frame feature complementarity.

Flow-guided aggregation employs optical flow to establish inter-frame feature correspondence \cite{guo2021frame}. Flow warps adjacent frame features to align with the current frame before aggregation \cite{muralidhara2022attention,han2020exploiting}. Compared to box-level aggregation, it enables flexible multi-frame fusion at earlier layers before final detection \cite{guo2021frame}. End-to-end learning can jointly optimize flow, features, and aggregation. Despite accuracy gains over single-frame detection, challenges include computational cost and handling large motions \cite{qian2020adaptive}.

Recent transformer-based architectures have shown promising results for advancing video object detection by enabling effective spatial-temporal reasoning. Notable approaches include TransVOD \cite{zhou2022transvod}, the first end-to-end transformer model for video detection without post-processing, and DETR \cite{carion2020end}, which eliminates hand-designed components in detectors via a transformer encoder-decoder architecture. Although originally for images, DETR has been adapted for video tasks \cite{he2021end}. Other examples are: TT-SRN \cite{wang2021end}, which aggregates spatial-temporal information for joint detection, segmentation, and tracking; and Swin Transformer \cite{liu2021swin}, a hierarchical model used for various vision tasks including video detection. TOD-Net \cite{cui2023feature} is another transformer-based framework improving query representations by feature aggregation.

Transformers and multi-frame feature aggregation offer complementary techniques for modeling temporal context. Transformer-based methods like TransVOD \cite{zhou2022transvod} and DETR \cite{he2021end} have achieved state-of-the-art accuracy by capturing long-range dependencies \cite{cui2023faq}. Multi-frame aggregation also boosts accuracy but is generally outperformed by transformers \cite{he2020temporal}. Multi-frame approaches have lower computational cost by operating on earlier features, while transformers are more expensive due to self-attention. However, efficient transformer designs are being explored \cite{cui2023faq}. Transformers inherently excel at temporal modeling through their architecture, whereas aggregation relies more on fixed schemes. Transformers also better handle varying video characteristics thanks to self-attention.

Recent works have combined both approaches, such as the FAQ method \cite{cui2023faq}, which aggregates inter-frame features to enhance transformer queries, and Attention-Guided Disentangled Feature Aggregation \cite{muralidhara2022attention}, a method that combines features from multiple frames to improve object detection by leveraging inter-frame correlations. These works show that combining transformers with multi-frame aggregation can improve accuracy by jointly modeling temporal context and leveraging inter-frame correlations. The integration of transformer self-attention and multi-frame feature aggregation remains an active area for advancing video understanding by utilizing their complementary strengths.

\subsubsection{3D Object Detection Approaches}
This section provides an overview of deep learning based 3D detection using different modalities and input representations, focusing on LiDAR and camera-based approaches as they are most prevalent.

\paragraph{LiDAR-Based 3D Object Detection}
LiDAR directly provides sparse 3D point clouds encoding precise geometric scene information. Earlier methods discretize the point cloud into 3D voxels and apply 3D convolutions. However, these are computationally expensive. More recent methods operate on raw point clouds by designing permutation invariant networks. PointNet \cite{gao2023pointpainting} is a pioneering work enabling direct point cloud processing. Subsequent works like PointRCNN \cite{shi2019pointrcnn}, Part-A2 Net \cite{shi2019part}, and PV-RCNN \cite{shreyas20213d} extend it for 3D detection by first generating proposals which are then refined using point features. Another line of work aggregates points into compact representations like pillars which encode vertical point columns, before applying efficient 2D convolutions on pseudo images. Pillar-based methods like PointPillars \cite{lang2019pointpillars} and PIXOR \cite{tian2022fully} are efficient but lose fine details. Recent pillar variants like SpindleNet \cite{zhao2017spindle} and CenterPoint \cite{yin2021center} improve representations by encoding local context more effectively. Range view methods like LaserNet utilize established image processing techniques by projecting point clouds into 2D range images and applying 2D convolutions.

\paragraph{Camera-Based 3D Object Detection}
Camera-based 3D object detection has been enhanced by the rich texture information provided by cameras. Earlier works, such as 3DOP \cite{kim20203d}, have lifted 2D detections into 3D using ground plane assumptions. More recent methods, including Mono3D \cite{chen2016monocular}, Mono3D++ \cite{he2019mono3d++}, and Pseudo-LiDAR \cite{you2019pseudo}, have improved performance by first estimating depth from images and then applying LiDAR-based detectors. Other approaches, such as Deep3DBox \cite{mousavian20173d} and ROI-10D \cite{manhardt2019roi}, have directly regressed 3D boxes from images without depth estimation, but have relied heavily on priors. Stereo cameras have provided better geometric constraints compared to monocular images, enabling techniques like Pseudo-LiDAR++ \cite{you2019pseudo} to further improve performance by combining depth and imagery.

\paragraph{Multi-Sensor Fusion}
LiDAR and camera provide complementary geometric and semantic information that can be fused to improve 3D detection accuracy, especially for small or distant objects \cite{liu2023real}. Early fusion methods like AVOD \cite{ku2018joint} fuse LiDAR and RGB early in the network, while late fusion approaches like Frustum PointNets \cite{qi2018frustum} use RGB to generate frustum proposals for LiDAR. PointPainting \cite{vora2020pointpainting} paints LiDAR points with semantic image features. PointFusion \cite{xu2018pointfusion} dynamically fuses multimodal features throughout the network. Recent techniques explore more robust fusion using attention mechanisms. The proposed TransFusion \cite{bai2022transfusion} method fuses LiDAR and images using a novel transformer architecture with soft-attention. It first generates initial boxes from LiDAR, then fuses image features in the second decoder layer using soft-attention. This provides robustness to misalignment and degraded image quality. Evaluated on KITTI and Waymo datasets, TransFusion outperforms prior fusion methods by 2-4\% in various metrics. It sets a new state-of-the-art for LiDAR-camera fusion in 3D detection by leveraging transformer attention to achieve soft-association between the modalities. The results validate that soft-attention based sensor fusion using transformers is an effective approach for handling misalignment and image degradations.

\subsection{Few-Shot Object Detection}
Few-shot object detection (FSOD) poses unique challenges compared to few-shot classification, as models need to accurately localize objects using extremely limited bounding box annotations. Several key techniques have been tailored specifically for few-shot object detection, including:

\begin{itemize}
    \item Two-stage detectors with incremental learning: In this approach, two-stage detectors like Faster R-CNN are modified for few-shot detection by incorporating separate classifier branches for novel classes. These branches are trained on the limited data, while keeping the weights of the base classes frozen during novel class training. This incremental learning strategy helps prevent interference \cite{belouadah2023multiod}.
    \item One-stage detectors with label smoothing: One-stage detectors such as YOLO are optimized for few-shot detection through the use of label smoothing techniques during training. By redistributing a portion of the target probability mass to non-ground truth classes, label smoothing improves model calibration and mitigates overfitting to the scarce training examples \cite{krothapalli2020one,lv2023detrs}.
    \item Transformer-based detectors: Transformer architectures, such as DETR, are particularly well-suited for few-shot detection. These architectures employ the self-attention mechanism to effectively model relationships between sparsely annotated examples. The absence of hand-designed components in transformers provides flexibility in adapting to few-shot detection scenarios \cite{jin2023incremental}.
    \item Advanced data augmentation: Specialized augmentation techniques, such as CutMix, play a crucial role in enhancing few-shot detection. CutMix involves blending object patches from different images to create new training examples, thereby improving the model's ability to handle few-shot scenarios. Additionally, techniques like class mixup and contextual augmentation help prevent overfitting in these challenging settings \cite{su2022dsla,liu2023recent}.
\end{itemize}

These techniques address the inherent difficulties faced in few-shot object detection, enabling models to overcome the limitations imposed by scarce annotations and perform object localization effectively. The main challenges in few-shot object detection include:

\begin{itemize}
    \item Localization from scarce bounding box annotations: Accurate localization becomes extremely difficult when there are only a few bounding box annotations available per novel class. The regression task becomes highly unstable, leading to poor generalizability. To address this researchers have explored techniques such as meta-learning \cite{jiang2023few} and instance weighting \cite{qiao2021defrcn} to improve localization performance with limited annotations.
    
    \item Imbalance between base and novel classes: Few-shot detection inevitably creates an imbalance between base classes with abundant training data and novel classes with only a few examples \cite{kohler2023few}. Focusing on frequent base classes often causes models to neglect rare new classes. This imbalance can be addressed using techniques like class balancing \cite{wu2021generalized}, where training samples from novel classes are augmented or weighted to alleviate the class imbalance problem.
    
    \item Domain shift between base and novel classes: Complex domain shifts often exist between the distributions of the base dataset and the novel classes that the model must adapt to \cite{jiaxu2021comparative}. Few-shot models struggle to transfer knowledge effectively across domains, leading to reduced performance on novel classes. To mitigate this challenge, researchers have proposed domain adaptation methods, such as domain alignment \cite{kang2019few} or domain generalization \cite{shangguan2023identification}, to align the feature distributions between the base and novel classes.
    
    \item Context modeling from limited examples: With only a few examples available, modeling contextual relationships between objects in a scene becomes highly ambiguous and uncertain \cite{liu2023recent}. This lack of context information makes few-shot detection unreliable. To overcome this challenge, researchers have explored techniques such as attention mechanisms \cite{wang2020frustratingly} and graph-based reasoning \cite{wang2022few} to incorporate context information and improve the detection performance of novel classes.
    
    \item Prevention of overfitting: Modern deep detection models with high capacity easily overfit to the scarce few-shot data, memorizing training examples without generalizing well to new instances \cite{kohler2023few}. This demands principled regularization techniques tailored for few-shot scenarios. Regularization techniques such as dropout, weight decay, and early stopping are commonly applied to prevent overfitting and improve the generalization ability of few-shot detection models \cite{liu2023recent}.
\end{itemize}

These challenges highlight the complexity of few-shot object detection and the need for innovative techniques to address them. Researchers are actively exploring novel algorithms and approaches to overcome these challenges and improve the performance of few-shot object detection models. For example, advanced data augmentation techniques have shown promise for synthesizing useful training signals from limited data. However, as helpfully pointed out in \cite{wang2020data}, there are information-theoretic limits on the additional `information' that can be fabricated from scarce examples. Recent studies such as \cite{franceschi2017theoretical} have analyzed these limits for few-shot learning, finding diminishing returns on augmentation beyond a certain point. Other works like \cite{cubuk2019randaugment} have proposed more principled augmentation approaches that consider information-theoretic measures to maximize diversity within feasible bounds. But further research is still needed into specialized augmentation techniques that work within information-theoretic constraints to provide useful signals without overstepping the true information content of limited training data. In addition to data augmentation, algorithms like \cite{vinyals2016matching,snell2017prototypical} aim to improve few-shot detection through other techniques such as meta-learning, metric-based learning, context modeling, and transfer learning. There are still many possibilities to enhance few-shot object detection through innovations in modeling, training strategies, evaluation protocols, and datasets. By overcoming challenges like scarce annotations, class imbalance, and domain shifts, researchers can achieve significant progress in few-shot detection and minimize dependency on extensive supervised data.

\clearpage

\section{Few-Shot Video Object Detection}
\begin{figure*}
    \centering
    \includegraphics[scale=0.1]{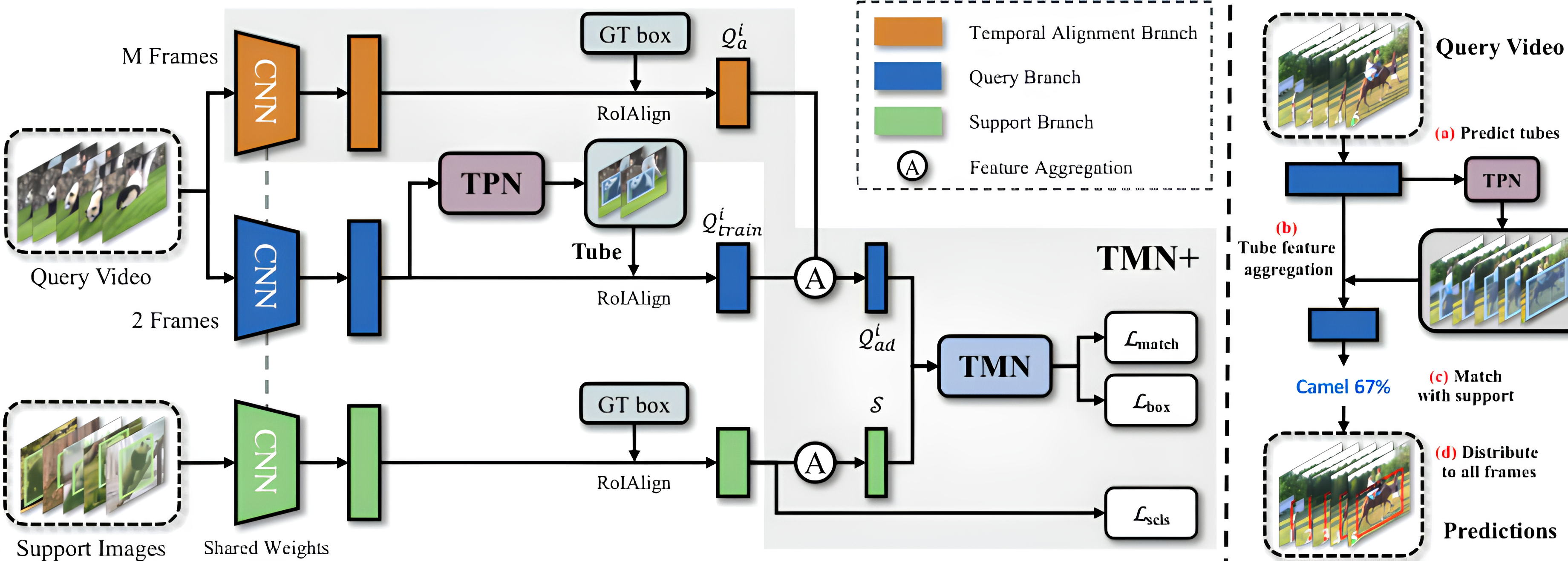}
    \caption{Overview of the two-stage architecture for few-shot video object detection proposed by \cite{fan2022few}. It consists of: 1) A Tube Proposal Network (TPN) that generates spatiotemporal proposals representing object trajectories across frames. 2) A Tube Matching Network (TMN) that classifies tube proposals by matching against few-shot support examples using multi-relation modules. The Temporal Alignment Branch (TAB) aligns query features across frames before matching. This design applies various techniques such as metric learning, temporal feature aggregation, and episodic training to enable effective few-shot detection in videos.}
    \label{fig:fs_vdo_1}
\end{figure*}

\begin{figure*}
    \centering
    \includegraphics[scale=0.07]{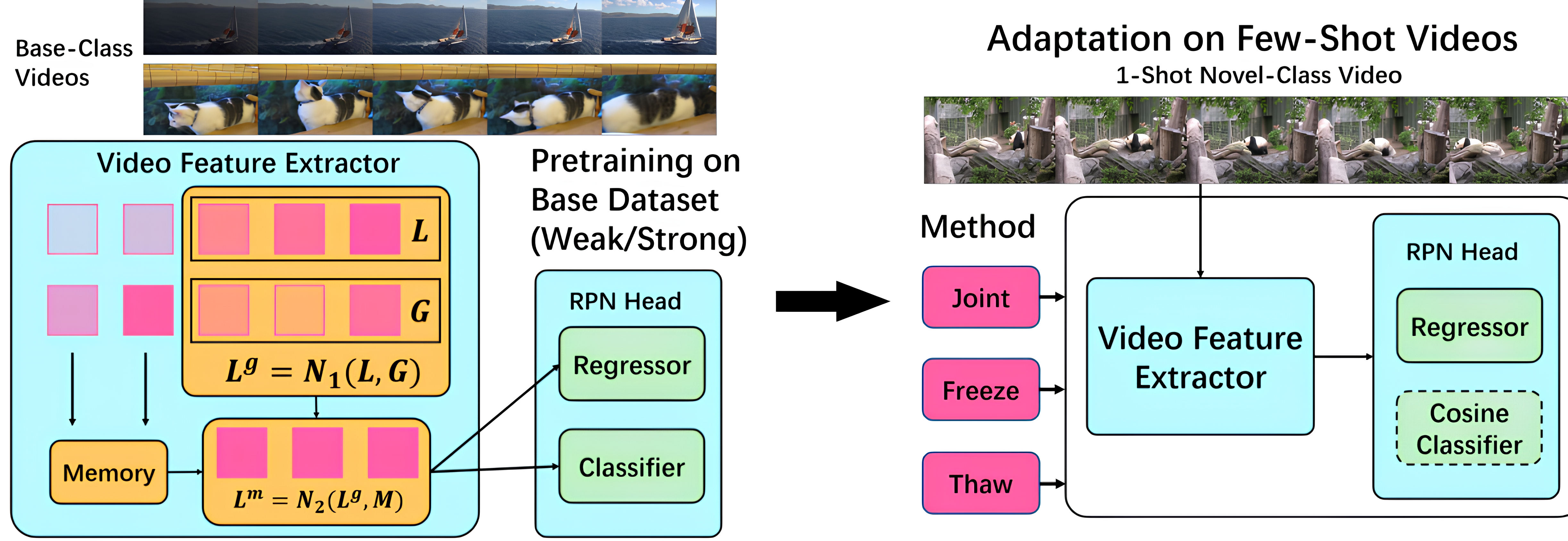}
    \caption{Overview of the Thaw architecture for few-shot video object detection proposed by \cite{yu2022few}. It consists of three key phases: 1) Pretraining on base classes using the MEGA model to extract multi-level local and global spatiotemporal features from input videos. 2) Adaptation on novel classes by adding a cosine classifier tuned on the scarce novel training data. 3) Fine-tuning using techniques like model freezing and gradual unfreezing to balance knowledge retention and adaptation. The tailored two-stage training process and integration of spatial-temporal information from video enables effective few-shot detection.}
    \label{fig:fs_vdo_2}
\end{figure*}

\begin{figure*}
    \centering
    \includegraphics[scale=0.08]{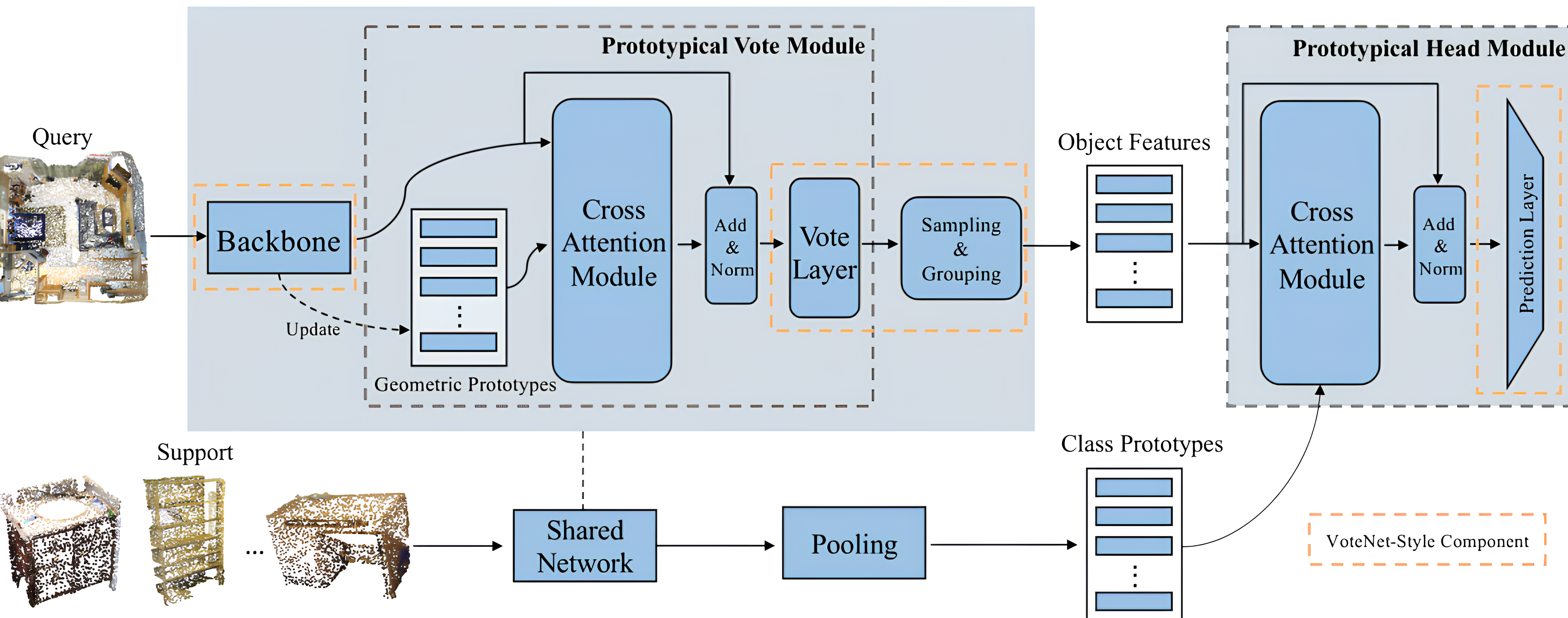}
    \caption{Overview of the Prototypical VoteNet architecture for few-shot 3D object detection \cite{zhao2022prototypical}. It contains two key components - the Prototypical Vote Module (PVM) which refines local features using geometric prototypes, and the Prototypical Head Module (PHM) which enhances object features using class-specific prototypes derived from the few-shot support examples. The class-agnostic PVM and class-specific PHM work together to enable effective FSL.}
    \label{fig:fs_3d_votenet}
\end{figure*}

\clearpage

\section{Few-Shot 3D Object Detection}

\begin{figure*}
    \centering
    \includegraphics[scale=0.068]{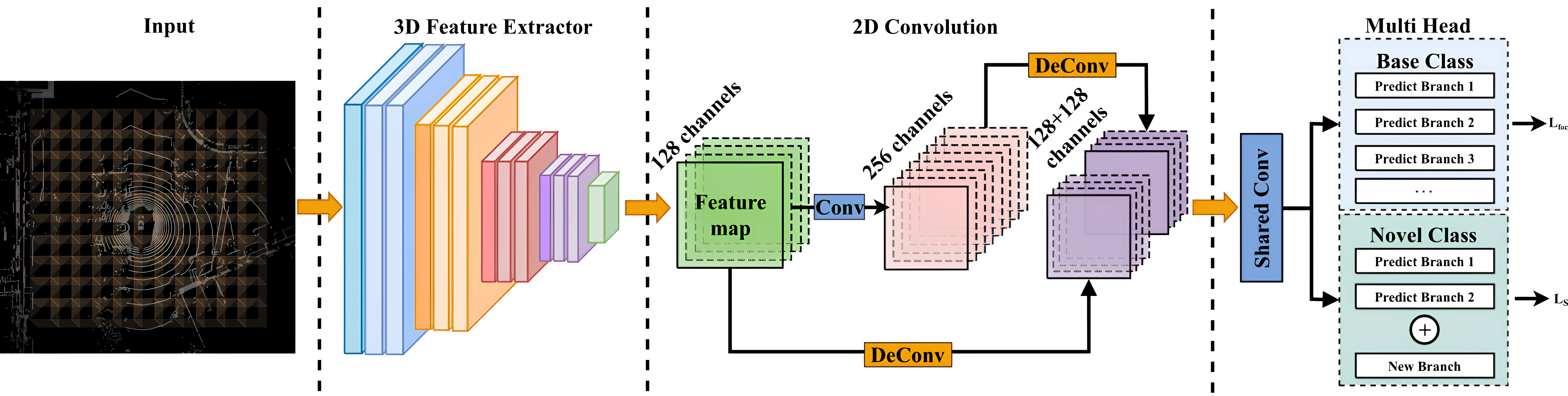}
    \caption{Overview of the generalized few-shot 3D object detection framework proposed by Liu et al. \cite{liu2023generalized}. It consists of a 3D feature extractor based on VoxelNet, followed by a region proposal network (RPN). A shared convolution layer feeds into separate prediction heads for base and novel classes, with incremental branches added for each novel class.}
    \label{fig:fs_3d_lidar1}
\end{figure*}

\begin{figure*}
    \centering
    \includegraphics[scale=0.08]{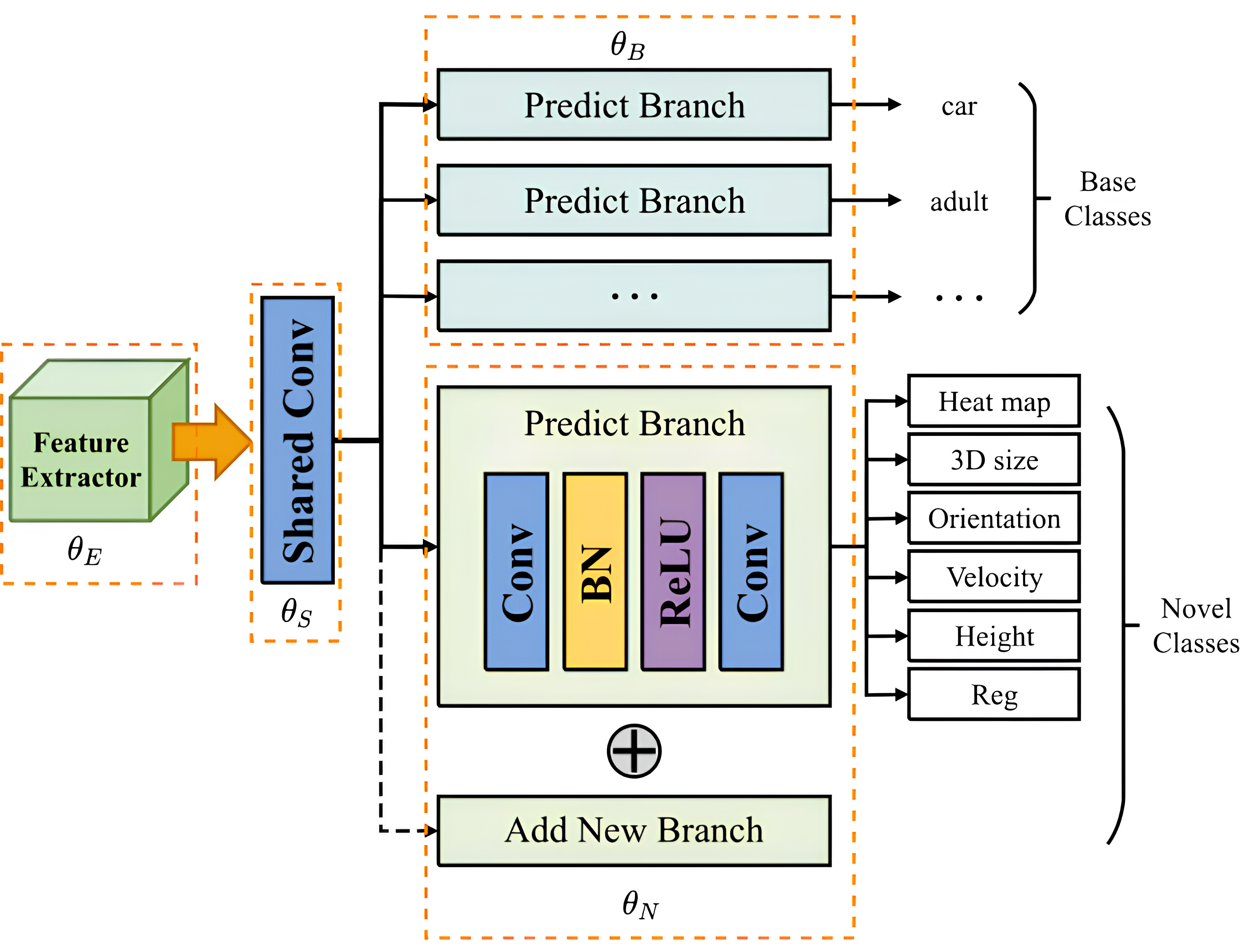}
    \caption{The incremental classifier branches tailored for each novel class in the few-shot 3D detection framework \cite{liu2023generalized}. The branches share an earlier convolutional layer with the base classes to avoid interference. Only the novel class branches are updated during the fine-tuning stage.}
    \label{fig:fs_3d_lidar2}
\end{figure*}

\begin{figure*}
    \centering
    \includegraphics[scale=0.09]{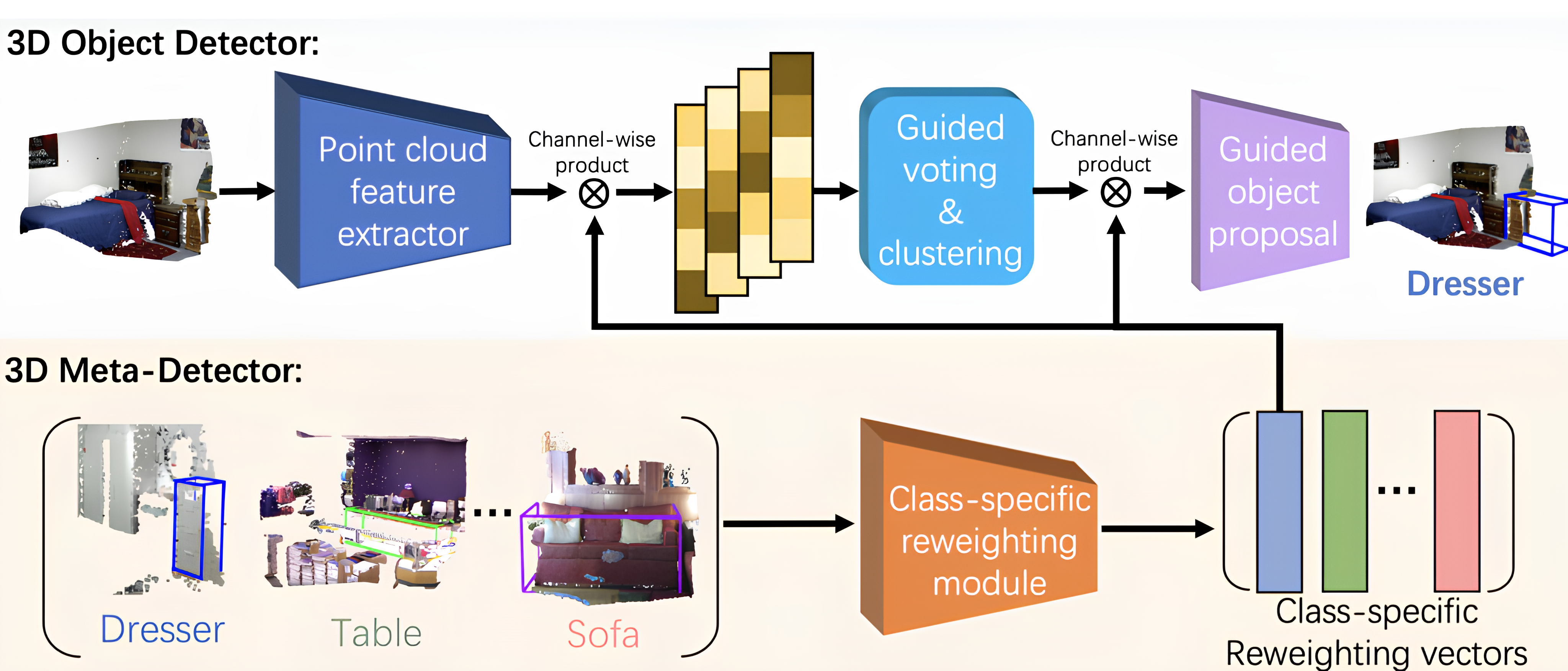}
    \caption{Overview of the MetaDet3D framework for few-shot 3D object detection \cite{yuan2022meta}. It consists of a 3D Meta-Detector module that generates class-specific reweighting vectors zn from the few-shot support points. These reweighting vectors guide the 3D Object Detector module, which contains point feature extraction, guided voting and clustering, and guided object proposal components. The reweighting vectors transfer knowledge from the scarce supports to enhance few-shot detection.}
    \label{fig:fs_3d_meta3ddet}
\end{figure*}

\begin{figure*}
    \centering
    \includegraphics[scale=0.1]{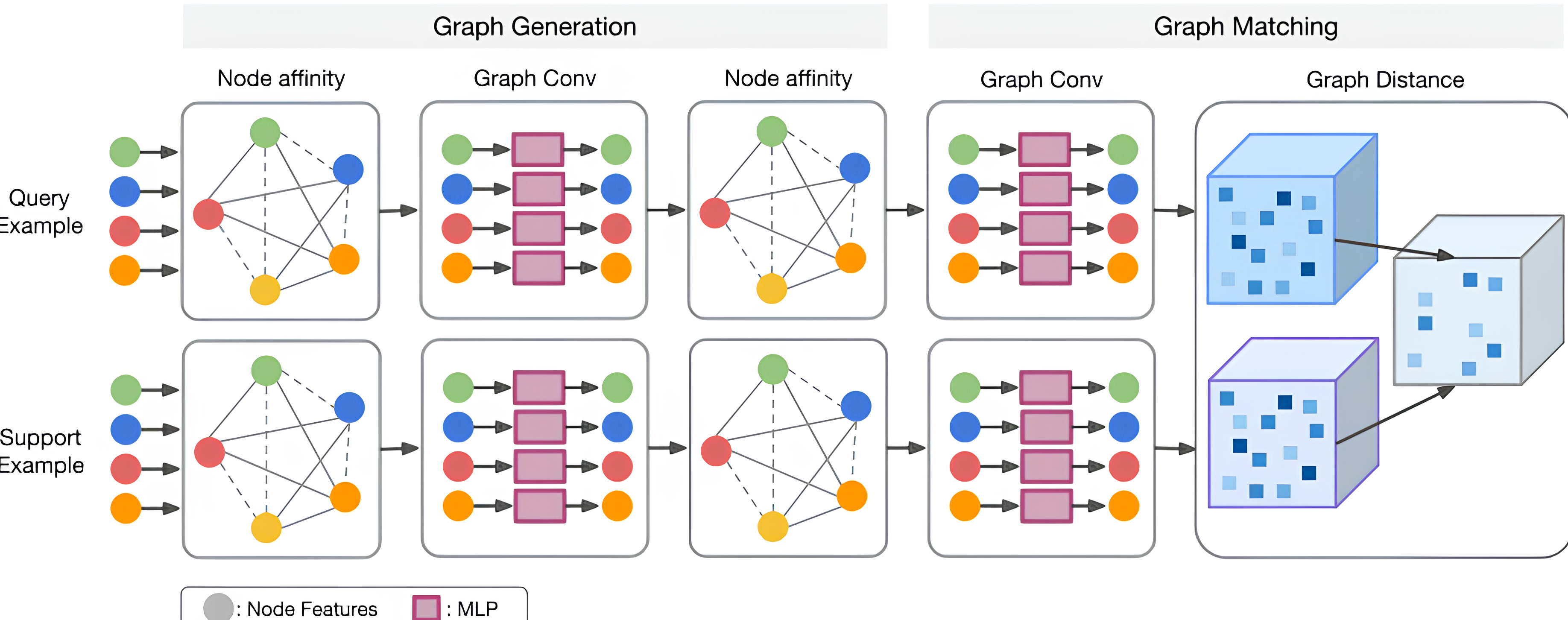}
    \caption{Overview of the Neural Graph Matching (NGM) Networks approach for few-shot 3D action recognition \cite{guo2018neural}. Videos are encoded into graph representations, with nodes as frame features and edges capturing temporal relationships. Graph matching is performed between support and query graphs by comparing node and edge features using cosine similarity. This structural matching enables effective FSL.}
    \label{fig:fs_3d_ngm}
\end{figure*}

\begin{figure*}
    \centering
    \includegraphics[scale=0.1]{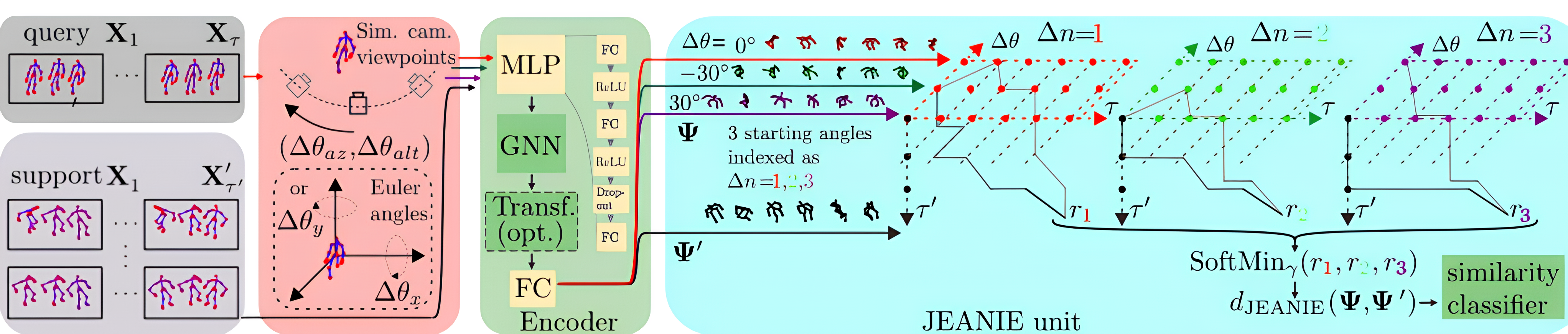}
    \caption{Overview of the few-shot action recognition framework on 3D skeletal sequences proposed by Wang et al. \cite{wang2022temporal}. It consists of two key components - the Encoding Network (EN) which models temporal dynamics from skeletal blocks, and the Joint tEmporal and cAmera viewpoiNt alIgnmEnt (JEANIE) module which aligns sequences in both time and viewpoint space for robust matching.}
    \label{fig:fs_3d_janie}
\end{figure*}

\end{document}